\pdfoutput=1
\documentclass[11pt]{article}

\usepackage[preprint]{acl}

\usepackage{times}
\usepackage{latexsym}

\usepackage[T1]{fontenc}
\usepackage[utf8]{inputenc}

\usepackage{microtype}
\usepackage{enumitem}
\usepackage{inconsolata}

\usepackage{graphicx}

\usepackage{hyperref}
\usepackage{url}
\usepackage[normalem]{ulem}
\usepackage{multirow}
\usepackage{booktabs}
\usepackage{colortbl}
\usepackage{booktabs} 
\usepackage{multirow} 
\usepackage{colortbl} 
\usepackage{rotating}
\usepackage[font=footnotesize]{caption}
\usepackage{graphicx}
\usepackage{multirow}
\usepackage{colortbl} 
\usepackage{graphicx}
\usepackage[table]{xcolor}  
\usepackage{array}
\usepackage{wrapfig}
\usepackage{xcolor, colortbl}
\usepackage{amssymb}
\usepackage{multirow}  
\usepackage{pifont}    
\usepackage{amsmath}   
\usepackage{xcolor}    
\usepackage{pifont}  
\usepackage{makecell}  
\usepackage[table,xcdraw]{xcolor}
\usepackage{amsmath}
\usepackage{graphicx}
\usepackage{array}
\usepackage[linesnumbered, ruled,vlined]{algorithm2e}

\usepackage[table]{xcolor}  
\definecolor{LightBlue}{RGB}{247, 252, 255}
\definecolor{myred}{RGB}{135,3,3}
\definecolor{red}{HTML}{c21313} % 
\newcommand{\uaa}[1]{\textcolor{red}{\scriptsize↑#1}}
\definecolor{LightRed}{RGB}{254, 248, 248}
\definecolor{darkgreen}{rgb}{0.0, 0.5, 0.0}

\usepackage{algorithmic}

\hypersetup{
    colorlinks=true,
    linkcolor=red,
    citecolor=cyan,
    filecolor=magenta,      
    urlcolor=cyan,
    }

\title{\textsc{Modest-Align}: Data-Efficient Alignment for Vision-Language Models}

\author{
 \textbf{Jiaxiang Liu\textsuperscript{1,2}}\ \ \
 \textbf{Yuan Wang\textsuperscript{2}}\ \ \
 \textbf{Jiawei Du\textsuperscript{3, 4}}\ \ \ \\
 \textbf{Joey Tianyi Zhou\textsuperscript{3, 4}}\ \ \
  \textbf{Mingkun Xu\textsuperscript{*, 1}}\ \ \
 \textbf{Zuozhu Liu\textsuperscript{*, 2}}
\\
\\
 \textsuperscript{1} \small Guangdong Institute of Intelligence Science and Technology, China\\
 \textsuperscript{2} \small ZJU-Angelalign R\&D Center for Intelligence Healthcare, Zhejiang University, China\\
 \textsuperscript{3} \small Centre for Frontier AI Research (CFAR), Agency for Science, Technology and Research (A*STAR), Singapore
\\
 \textsuperscript{4} \small Institute of High Performance Computing (IHPC), Agency for Science, Technology and Research (A*STAR), Singapore
\\
 { \small
   {\tt \{forworkliu\}@gmail.com}
 }
}

\begin{document}
\maketitle
\begin{abstract}
Cross-modal alignment aims to map heterogeneous modalities into a shared latent space, as exemplified by models like CLIP, which benefit from large-scale image-text pretraining for strong recognition capabilities.
However, when operating in resource-constrained settings with limited or low-quality data, these models often suffer from overconfidence and degraded performance due to the prevalence of ambiguous or weakly correlated image-text pairs. Current contrastive learning approaches, which rely on single positive pairs, further exacerbate this issue by reinforcing overconfidence on uncertain samples.
To address these challenges, we propose \textsc{Modest-Align}, a lightweight alignment framework designed for robustness and efficiency. Our approach leverages two complementary strategies—\textit{Random Perturbation}, which introduces controlled noise to simulate uncertainty, and \textit{Embedding Smoothing}, which calibrates similarity distributions in the embedding space. These mechanisms collectively reduce overconfidence and improve performance on noisy or weakly aligned samples.
Extensive experiments across multiple benchmark datasets demonstrate that \textsc{Modest-Align} outperforms state-of-the-art methods in retrieval tasks, achieving competitive results with over 100× less training data and 600× less GPU time than CLIP. Our method offers a practical and scalable solution for cross-modal alignment in real-world, low-resource scenarios. 
% Code will be publicly released.
\end{abstract}

\begin{figure}[ht!]
\centering
% \label{logits_dataset}
\includegraphics[width=0.5\textwidth]{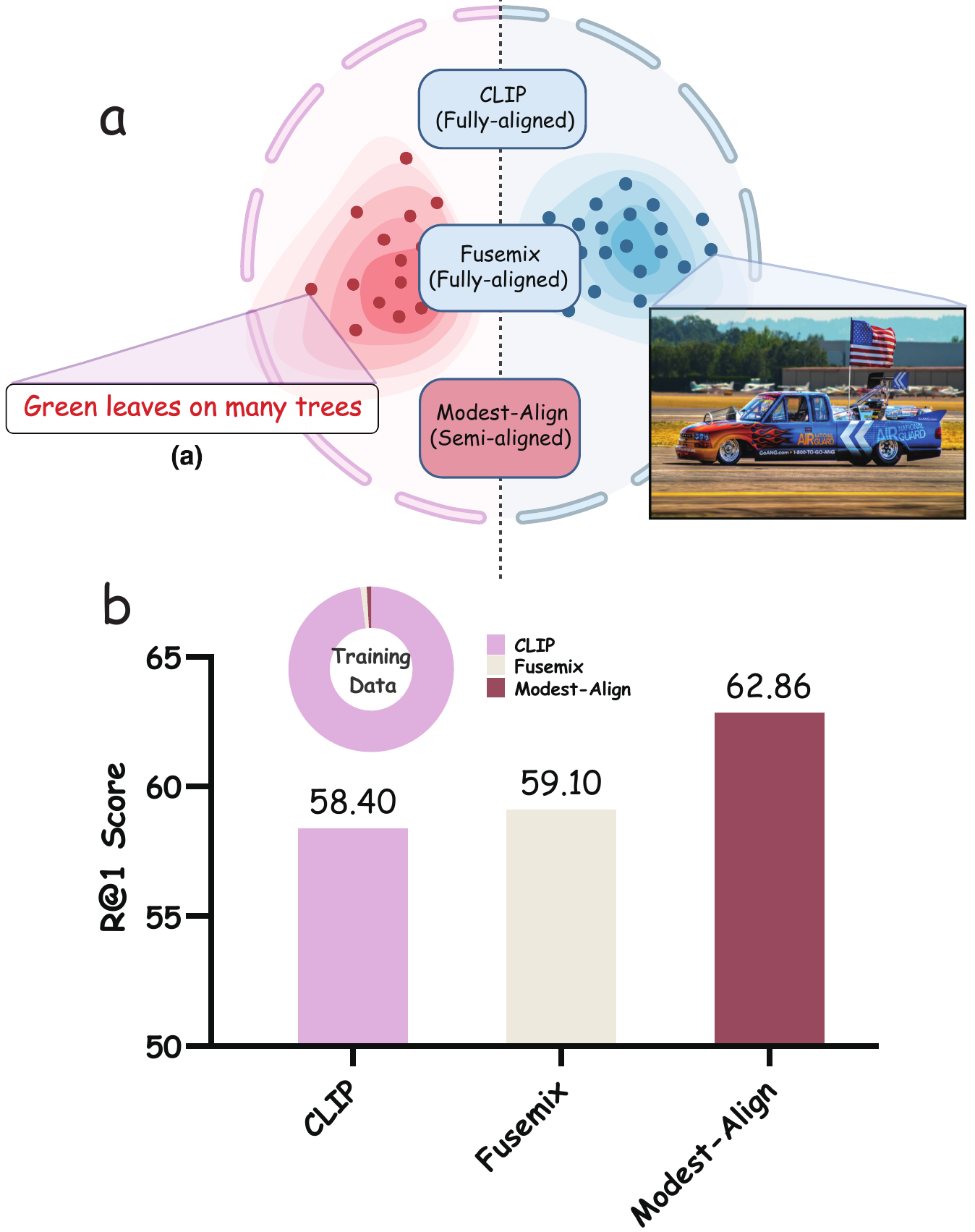}
\caption{
(a) Compared with CLIP and Fusemix, which assume full alignment during training (e.g., “Green leaves on many trees” is treated as a fully matched pair), our Modest-Align handles partially matched or noisy positive pairs by encouraging semi-aligned representations through tailored perturbation and smoothing strategies.
(b) On the MS-COCO test dataset (Image → Text retrieval), Modest-Align outperforms both CLIP and Fusemix, despite using only 3.5M training samples—exceeding Fusemix trained on 5M samples (including our 3.5M subset) and CLIP trained on 400M pairs.
} 
\label{new_emnlpFigure}
\vspace{-1em}
\end{figure}

\begin{figure*}[htbp]
\centering
% \label{logits_dataset}
\includegraphics[width=\textwidth]{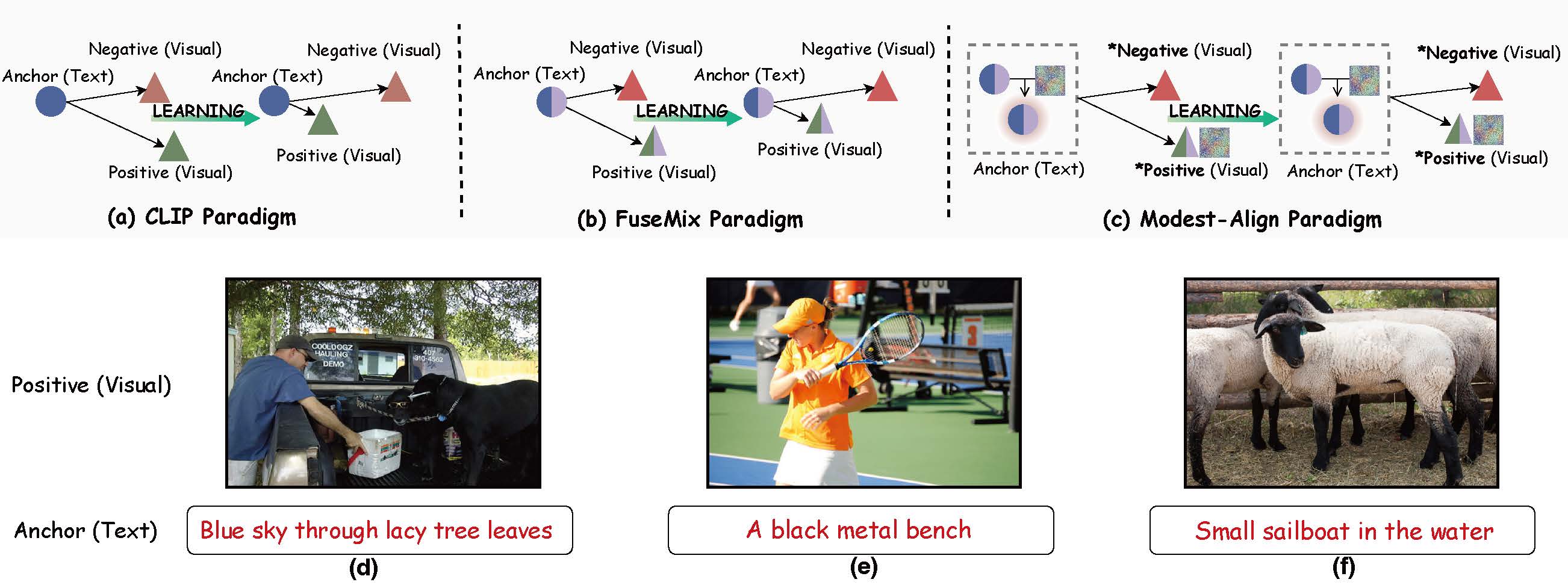}
\caption{
(a-c) Three Paradigms for Cross-modal Alignment: CLIP for contrastive learning, FuseMix for embedding-level mixing, and \textsc{Modest-Align} for moderating overconfidence and enhancing robustness.
Both FuseMix and \textsc{Modest-Align} derive positive pairs from mixed features, while \textsc{Modest-Align} further injects random perturbations into visual and textual embeddings to simulate uncertainty.
“Positive” and “Negative” denote matched and mismatched pairs, with the gray region indicating uncertainty-aware perturbation.
(d–f) Examples of ambiguous samples, including partially matched pairs (d, e) and completely mismatched pairs (f).
} 
\label{Figure_3paradigms}
\vspace{-1em}
\end{figure*}

\section{Introduction}

Multimodal learning, by integrating different types of data modalities, enhances a model's perception and understanding capabilities, facilitating cross-modal information interaction and integration \cite{radford2021clip,vouitsis2024data,zhai2022lit,liu2023parameter,liu2023deep,yang2021infrared,girdhar2023imagebind,liu2024vpl,liu2024medcot}. 
Recent advancements in multimodal machine learning have shown unprecedented potential across various application fields, with some applications even attracting mainstream attention \cite{girdhar2023imagebind,radford2021clip}. 
The cornerstone of multimodal learning is cross-modal alignment, which maps information from multiple modalities, such as text and images, into a unified multimodal vector space \cite{radford2021clip,alayrac2022flamingo}.
Researchers have made numerous efforts in cross-modal alignment, with Visual Language Models (VLMs) being particularly representative. 
VLMs like CLIP \cite{radford2021clip}, which undergo extensive image-text pre-training, excel in image recognition tasks, showcasing the potential of VLMs in establishing effective cross-modal connections.

The success of cross-modal alignment largely relies on large-scale training mechanisms like CLIP, which often require extensive GPU resources and rely on billions of multimodal data pairs \cite{zhai2022lit,radford2021clip,alayrac2022flamingo}.  
However, the high computational costs are impractical for scenarios with limited computing resources or scarce multimodal data. 
Therefore, designing a cost-effective and efficient cross-modal alignment framework is crucial. 
Inspired by Mixup \cite{zhang2018mixup}, Fusemix introduces an efficient strategy for cross-modal alignment \cite{vouitsis2024data} by augmenting the latent spaces of pre-trained unimodal encoders, allowing for model creation with significantly reduced data and computational requirements.
However, ambiguous samples—whether partially matched or completely unmatched—in datasets with weakly associated (see \autoref{new_emnlpFigure}), low-quality image-text pairs can lead to overconfidence and confusion in models, ultimately degrading performance. 
Moreover, current contrastive learning methods, which rely on single positive examples, exacerbate this issue by further encouraging overconfidence in the presence of ambiguous samples \cite{vouitsis2024data}.

To overcome these issues, we propose \textsc{Modest-Align}, a cross-modal alignment enhancement method designed to adjust the matching degree of the
data and moderate model overconfidence with a single GPU, incorporating two key components:
1) \textit{Random Perturbation}: This introduces normally distributed perturbations at the visual-text feature level to simulate uncertainty, enhancing the model's generalization capability and helping it learn more robust feature representations.
2) \textit{Embedding Smoothing}: This aims to smooth the model’s prediction of output distributions, moderating model overconfidence in positive samples and increasing the smoothness for predictions on uncertain samples, thereby enhancing generalization.
By using \textsc{Modest-Align} to align the latent spaces of pre-trained unimodal encoders, we have developed a highly competitive visual-language (V-L) model. In retrieval tasks, this model not only surpasses existing state-of-the-art (SoTA) methods but also significantly reduces the need for computational resources and data, as detailed in \autoref{Figure_1}.
Our study makes two significant contributions:
\begin{itemize}
 \item  
 % We discovered that the quality of existing image-text datasets is suboptimal, leading to visual-language (V-L) models becoming confused and overly confident when faced with ambiguous positive pairs. This significantly impacts model performance.
 Theoretical and empirical analysis reveals that the quality of existing image-text paired datasets is suboptimal, causing VLMs to become confused and overly confident when faced with ambiguous positive pairs (either partially matched (\autoref{Figure_3paradigms} \textcolor{red}{d}, \textcolor{red}{e}) or completely unmatched (\autoref{Figure_3paradigms} \textcolor{red}{f}). This significantly undermines cross-modal alignment.

 \item  
 We propose a novel V-L alignment method, \textsc{Modest-Align}, which incorporates \textit{Random Perturbation} to simulate input uncertainty and \textit{Embedding Smoothing} to mitigate overconfidence in positive samples. This \textsc{Modest-Align} enhances model generalization and robustness, effectively addressing the challenges posed by the suboptimal quality of existing datasets.

\end{itemize}

\begin{figure*}[htbp]
\centering
% \label{logits_dataset}
\includegraphics[width=\textwidth]{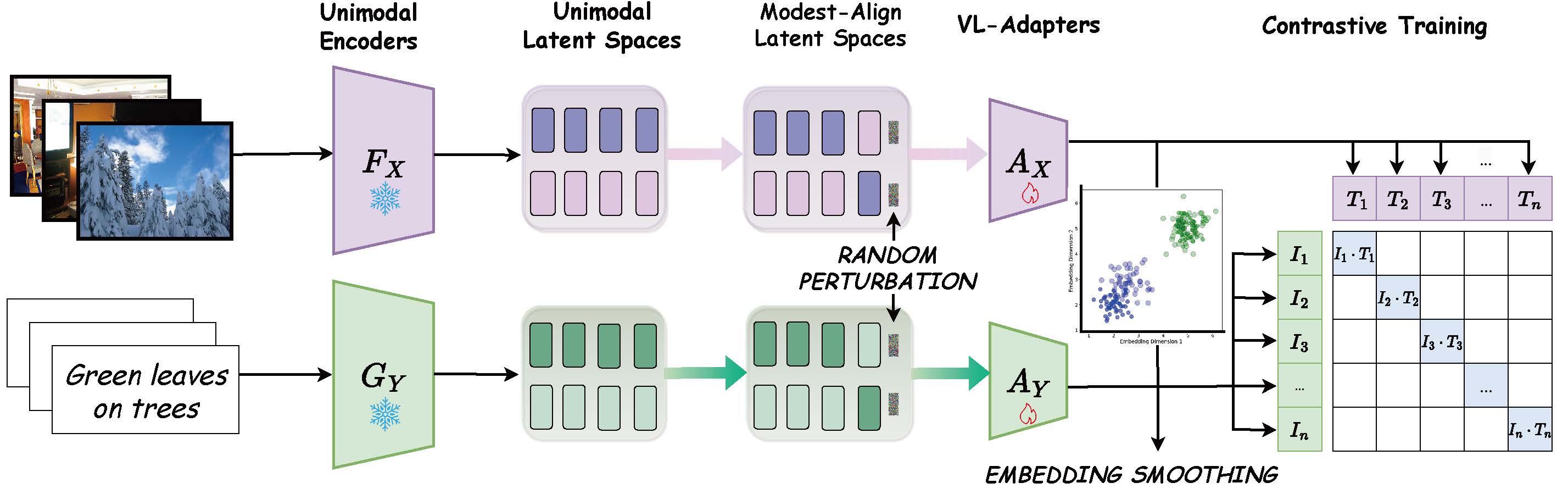}
\caption{
A pipeline of the \textsc{Modest-Align} showcases the process of aligning the latent spaces of pre-trained unimodal encoders using a fewer dataset of paired data. 
The unimodal encoders remain frozen, with their latent encodings pre-computed only once for efficiency. In this framework, both Random Perturbation and Embedding Smoothing are applied to each latent space to enhance robustness and reduce model overconfidence. Lightweight V-L adapters are trained to meticulously align these augmented latents into a cohesive, shared latent space, effectively bridging the semantic gap between different modalities.
} 
\label{Figure_pipeline}
\vspace{-1em}
\end{figure*}

\section{Related Work}
\label{sec: related work}

Cross-modal alignment achieves cross-modal synchronization not through direct correspondences between modalities, but implicitly via internal model mechanisms that discern latent semantic connections within the data. The primary objective of these models is to learn a shared latent space capable of jointly encoding multiple modalities, thereby facilitating effective cross-modal alignment \cite{tan2019lxmert, li2020oscar, yuan2021florence, wang2022internvideo, bao2022vlmo, wang2022simvlm, girdhar2022omnivore, likhosherstov2023polyvit, zhang2023metatransformer, wu2023next}.
Image-language alignment is a pivotal area of study in cross-modal alignment, aiming to create universal models capable of interpreting both image and language data. 
Standard multimodal models usually undergo end-to-end training on image-text pairs. Yet, training these large-scale models from scratch demands substantial computational and data resources, which can restrict scalability.
\cite{arandjelovic2017looklistenlearn, lu2019vilbert, sun2019videobert, su2020vlbert, chen2020uniter, li2021albef, li2022blip,liu2025kpl}.
% A more practical approach leverages guidance from pretrained unimodal networks. Yet, these efforts often still require backpropagation through the pretrained networks, which is computationally expensive due to the large scale of unimodal networks, and this issue intensifies as network size increases.

Pioneered by CLIP and ALIGN \cite{radford2021clip,jia2021align}, this approach uses a dual-encoder architecture, jointly embedding text and images into the same latent space through contrastive target training. 
% CoCa introduces an autoregressive image captioning element to contrastive targets, enhancing model performance. 
3T aligns text and image encoders with the latent space of a pretrained classifier \cite{kossen2023three}. 
LiT uses a frozen pretrained image classifier as the image encoder and aligns a text encoder to it \cite{zhai2022lit}.
Although these methods have seen success, they mostly train one or two encoders from scratch, relying on expensive cross-GPU gradient computations. 
ImageBind \cite{girdhar2023imagebind} uses images as anchors to learn a shared latent space across six modalities through contrastive learning, jointly training various modality encoders from scratch. 
Moreover, the large-scale image-text paired datasets they use, ranging from 400 million to 5 billion pairs, mostly sourced from the internet, are generally not public \cite{vouitsis2024data}.
In contrast to these works, Fusemix boosts computational and data efficiency through feature augmentation techniques, using frozen pre-trained unimodal encoders and fewer multimodal paired data, requiring fewer resources \cite{vouitsis2024data}. 
However, ambiguous positive samples in weakly associated datasets (see \autoref{new_emnlpFigure}) lead to model overconfidence and degraded performance, exacerbated by contrastive learning methods that focus on single positive examples \cite{radford2021clip}.
% In contrast to these works, Fusemix computational and data efficiency by using frozen pre-trained unimodal encoders, by leveraging minimal multimodal paired data, and by ensuring all our experiments require .  
% Fusemix insight is that unimodal encoders, already pretrained on extensive unimodal data, encode rich semantic information that can effectively guide multimodal fusion. 

\autoref{Figure_3paradigms} compares three alignment paradigms: the CLIP Paradigm, which utilizes contrastive learning to manage data point relationships; the Fusemix Paradigm, which enhances embeddings by mixing features of image-text pairs; and the \textsc{Modest-Align} Paradigm, which reduces overconfidence through perturbations and embedding smoothing, enhancing generalization and robustness. The \textsc{Modest-Align} specifically reduces model overconfidence and ensures experiments are computationally and data efficient, requiring only a reasonable amount of GPU resources (\autoref{Figure_days}).

\section{Methodology}
\label{method}

In this section, we introduce the \textsc{Modest-Align} framework, designed to facilitate visual-text modal alignment in the latent space while addressing key considerations such as model overconfidence, and computational and data efficiency.
\textsc{Modest-Align} entire process is illustrated in \autoref{Figure_pipeline}.

\subsection{Preliminaries}
\label{headings}

\textbf{Notation:}
We define the task of V-L alignment from an alignment perspective. 
The goal is to learn a shared latent space between visual and textual modal inputs. 
Formally, given any two data modalities (images \( \mathcal{X} \) and text \( \mathcal{Y} \)), 
our objective is to learn two networks, \( f_X: \mathcal{X} \rightarrow \mathcal{S} \) and \( f_Y: \mathcal{Y} \rightarrow \mathcal{S} \), that embed each modality into a shared latent space \( \mathcal{S} \).

We take our two encoders as $f_X=F_X \circ A_X$  and $f_Y=G_Y \circ A_Y$. 
That is, we define $F_X (Frozen) \colon \mathcal{X} \to \mathcal{U_X}$ and $G_Y (Frozen) \colon \mathcal{Y} \to \mathcal{U_Y}$, where $\mathcal{U_X}$ and $\mathcal{U_Y}$ are intermediate latent spaces.
We then have $A_X (Learnable) \colon \mathcal{U_X} \to \mathcal{S}$ and $A_Y (Learnable) \colon \mathcal{U_Y} \to \mathcal{S}$, 
which we hereafter refer to as V-L adapters. 
Our insight here is to take both $F_X$ and $G_Y$ as pre-trained unimodal encoders which we keep frozen throughout, and treat our V-L adapters $A_X$ and $A_Y$ as learnable heads for cross-modal alignment. 
Therefore, we can define our learning objective using the InfoNCE loss function as follows:

\begin{equation}
\small
\mathcal{L}_{\text{NCE}} = - \log \frac{\exp(\text{sim}(f_X(\mathcal{X}_i), f_Y(\mathcal{Y}_i)) / \tau)}{\sum_{j=1}^{N} \exp(\text{sim}(f_X(\mathcal{X}_i), f_Y(\mathcal{Y}_j)) / \tau)},
\end{equation}
\normalsize
where $\text{sim}(\cdot, \cdot)$ denotes the similarity function, $\tau$ is the temperature parameter, and $N$ is the number of all samples. Where \( \mathcal{X}_i \) and \( \mathcal{Y}_i \) are positive pairs.

Although Eq.~(1) provides a standard contrastive formulation to align visual and textual modalities via $f_X = F_X \circ A_X$ and $f_Y = G_Y \circ A_Y$, it assumes sufficient training samples to balance noise and outliers. However, in low-resource scenarios where $N \ll 400\!M$, many positive pairs $(\mathcal{X}_i, \mathcal{Y}_i)$ may be weakly correlated or noisy, which can lead to overfitting or misalignment during training.

\noindent\textbf{Problem Formulation:} To explicitly promote alignment efficiency under limited data, we reformulate the objective from a \textit{unit sample information efficiency} perspective. Specifically, we propose to maximize the alignment quality \emph{per training pair} as follows:

\begin{equation}
\max_{\theta} \;\frac{1}{N} \sum_{i=1}^{N} \mathcal{L}_{\text{align}}\left( f_X(\mathcal{X}_i), f_Y(\mathcal{Y}_i) \right),
\tag{2}
\end{equation}

where $f_X(\mathcal{X}_i)$ and $f_Y(\mathcal{Y}_i)$ are the projected embeddings in the shared space $\mathcal{S}$ obtained through the visual and textual adapters $A_X$ and $A_Y$, respectively. $\mathcal{L}_{\text{align}}(\cdot, \cdot)$ measures the quality of alignment between the two modalities—e.g., cosine similarity or contrastive matching score.

Compared to batch-wise InfoNCE, Eq.~(2) focuses on {per–pair alignment efficiency}.  
Yet, under limited or noisy supervision this objective still inherits an
implicit \emph{hard–match} assumption: every designated positive pair
$(\mathcal{X}_i,\mathcal{Y}_i)$ is treated as \emph{perfectly} aligned,
while all other pairs in the mini-batch are considered negatives.

\paragraph{Alignment Distribution.}
% Let
% \[
% s_{ij}\;=\;\frac{\langle f_X(\mathcal{X}_i),\,f_Y(\mathcal{Y}_j)\rangle}{\tau},
% \]
% \[
% p_{ij}\;=\;\frac{\exp(s_{ij})}{\sum_{k=1}^{N}\exp(s_{ik})}\; ,
% \]
% Let 
% \[
% s_{ij} = \frac{\langle f_X(\mathcal{X}_i),\,f_Y(\mathcal{Y}_j)\rangle}{\tau},
% \]
% be the scaled similarity score between the $i$-th image and the $j$-th text embedding, where $\tau$ is a temperature parameter.

% Then, the predicted alignment distribution over text samples for a given image $\mathcal{X}_i$ is defined as:
% \[
% p_{ij} = \frac{\exp(s_{ij})}{\sum_{k=1}^{N}\exp(s_{ik})},
% \]
% where $p_{ij}$ denotes the probability of $\mathcal{X}_i$ being aligned with $\mathcal{Y}_j$.
Let 
\[
s_{ij} = \frac{\langle f_X(\mathcal{X}_i), f_Y(\mathcal{Y}_j)\rangle}{\tau}
\]
denote the scaled similarity between the $i$-th image and the $j$-th text embedding. The predicted alignment distribution is then:
\[
p_{ij} = \frac{\exp(s_{ij})}{\sum_{k=1}^{N}\exp(s_{ik})},
\tag{3}
\]
where $p_{ij}$ represents the normalized matching probability in matching $\mathcal{X}_i$ with $\mathcal{Y}_j$.

\paragraph{Failure modes under limited or noisy supervision.}
With the hard pairwise target
\(q_{ij}= \mathbf 1_{\{j=i\}}\),
the predicted distribution and its per-sample gradients are:
% \qquad
\[g_{ij}\;=\;\frac{\partial\mathcal L_i}{\partial s_{ij}}
           =\begin{cases}
              p_{ij}-1, & j=i,\\[2pt]
              p_{ij},   & j\neq i .
            \end{cases}
\tag{4}
\]

When the declared positive pair \((\mathcal X_i,\mathcal Y_i)\) is only
weakly aligned, the contrastive objective exhibits three coupled
pathologies:

\begin{enumerate}[label=(\alph*)]
\item \textbf{Over-emphasis on few pairs:}\;
      \(p_{ii}\!\ll\!1\) still triggers \(\max_j p_{ij}\!\to\!1\);
      the model forces \(s_{ii}\) upward and suppresses all
      \(s_{ij}\,(j\!\neq\! i)\).

\item \textbf{Entropy collapse:}\;
      the prediction entropy
      \[
      H(p_i)= -\sum_{j=1}^{N} p_{ij}\log p_{ij}\;\longrightarrow\;0,
      \]
      yielding an over-confident, brittle alignment distribution.

\item \textbf{Unstable gradients / embedding collapse:}\;
      the $\ell_1$-norm of the gradient vector
      \[
      \bigl\lVert\mathbf g_i\bigr\rVert_1
      = |p_{ii}-1|+\!\!\sum_{j\neq i} p_{ij}
      \;\approx\;1
      \]
      becomes large when \(p_{ii}\) is small,
      resulting in abrupt updates and eventual representation collapse.
\end{enumerate}

\subsection{Proposed Solution}

The large gradient norm in Eq.~(4) is rooted in two factors:
(i) the \emph{rigid} supervision $q_{ij}$ that drives $|p_{ii}-q_{ii}|
 = 1-p_{ii}$ to its maximum, and
(ii) the \emph{sharp} score $s_{ii}$ that contracts the probability
mass into a single point.
Therefore, an effective remedy must \textbf{(a)} soften the supervision
signal and \textbf{(b)} disperse the score distribution.
We next show that both requirements can be satisfied by a
\emph{single mathematical modification} of the optimization objective.

\paragraph{Random Perturbation (RP) as Jacobian regularization.}
We inject an \emph{input–space Gaussian perturbation} to \textit{both}
visual and textual embeddings; for the visual branch we have
\[
\tilde f_X(\mathcal X_i)=f_X(\mathcal X_i)+\sigma\epsilon_i,
\qquad
\epsilon_i\sim\mathcal N(0,I),
\]
and the textual branch is treated in the same manner. The perturbed similarity is defined as 
$\tilde s_{ij}=\langle\tilde f_X(\mathcal X_i),f_Y(\mathcal Y_j)\rangle/\tau$.
Taking expectation over $\epsilon_i$ we have, by a
first–order Taylor expansion,
\[
\mathbb E_{\epsilon}\!\bigl[\mathcal L_i(\tilde s_{i:})\bigr]
\;\approx\;
\mathcal L_i(s_{i:})
+\frac{\sigma^{2}}{2}\,\bigl\|\nabla_{\!\tilde f_X} 
\mathcal L_i\bigr\|^{2}.
\tag{5}
\]
which is equivalent to adding a \emph{Jacobian norm penalty} to the
original loss.
Consequently,
\[
\bigl\|\nabla_{\!\tilde f_X} 
\mathcal L_i\bigr\|_1
\;\uparrow\; \text{large}
\quad\Longrightarrow\quad
\mathcal L_{\text{RP}}\; \text{increases},
\]
forcing the optimiser to seek solutions with smaller gradients.
Equation~(5) reveals that perturbation training is \emph{equivalent} to
adding a Jacobian‐norm penalty, directly shrinking the gradient norm in
Eq.~(4) and hence preventing representation collapse.
For inference, the noise is removed to ensure accurate predictions, returning embeddings to their original state.

The large gradient norm in Eq.~(4) originates not only from the narrow
score landscape but also from the \emph{rigid} pairwise target
$q_{ij}$.  
This observation motivates a complementary remedy that operates directly on the \emph{target space}:

\paragraph{Embedding Smoothing (ES) as target relaxation.}
We replace the binary supervision by its convex relaxation:  
\[
\tilde q_{ij}
=(1-\alpha)\,\mathbf 1_{\{j=i\}}
+\frac{\alpha}{N}\,\mathbf 1,
\qquad 0<\alpha<1,
\tag{6}
\]
i.e.\ each sample retains weight \(1-\alpha\) on its annotated match and
shares the remaining mass \(\alpha\) uniformly with all other pairs.
The per‑pair loss becomes
\(
\mathcal L_i^{\text{ES}}
=\operatorname{KL}(\tilde q_i\,\|\,p_i).
\)
ES offers at least three key benefits:

\begin{itemize}
\item \emph{Gradient bound.}  
      In Eq.~(4) the critical term is now
      \(
      |p_{ii}-\tilde q_{ii}|
      =
      \bigl|p_{ii}-\,(1-\alpha+\tfrac{\alpha}{N})\bigr|
      \le 1-\alpha+\tfrac{\alpha}{N}
      < 1 ,
      \)
      providing an explicit upper limit on the update magnitude.
\item \emph{Entropy floor.}  
      Jensen’s inequality gives
      \(
      H(p_i)\;\ge\;-\!\bigl[(1-\alpha)\log(1-\alpha)+\alpha\log(\tfrac{\alpha}{N})\bigr],
      \)
      guaranteeing a non‑degenerate probability distribution and
      eliminating over‑confidence.
\item \emph{Convex relaxation.}  
      Because $\tilde q_i$ is a convex combination of a delta
      distribution and the uniform distribution, the KL objective
      remains convex in $p_i$, yielding a better behaved optimization
      landscape.
\end{itemize}

\paragraph{Confidence-Calibrated Contrastive Loss (CCL).}
Substituting the perturbed embeddings and smoothed targets into
Eq.~(2) yields the data-efficient alignment objective (CCL):
\[
\mathcal L_{\text{ccl}}
= \mathbb E_{\epsilon}\!\Bigl[
  \tfrac1N\!\!\sum_{i=1}^{N}
  \mathrm{KL}\!\bigl(
    \tilde q_i \,\|\,\mathrm{softmax}(\tilde s_{i:})
  \bigr)
\Bigr]
+\lambda\,\sigma^{2},
\tag{7}
\]
where $\sigma$ is the perturbation scale and $\lambda$ is the regularization coefficient.
Eq.~(7) unifies \textit{Embedding Perturbation} and
\textit{Embedding Smoothing} as a principled solution to the
instability identified in Eq.~(4). The pseudo-code is summarised in Algorithm~\ref{alg:modest}.
To our knowledge, this is the first formulation that explicitly integrates uncertainty modeling into a unified objective for cross-modal alignment.

\begin{algorithm}[t]
\caption{Training \textsc{Modest-Align}}
\label{alg:modest}
\begin{algorithmic}[1]
  \REQUIRE mini-batch $\{(\mathcal X_i,\mathcal Y_i)\}_{i=1}^{N}$

  \STATE $\mathbf z_x \leftarrow F_X(\mathcal X)$; \,
         $\mathbf z_y \leftarrow G_Y(\mathcal Y)$
         \hfill

  \STATE sample $\epsilon,\epsilon' \sim \mathcal N(\mathbf 0,I)$

  \STATE $\tilde{\mathbf z}_x \leftarrow \mathbf z_x + \sigma\epsilon$; \,
         $\tilde{\mathbf z}_y \leftarrow \mathbf z_y + \sigma\epsilon'$
         \hfill

  \STATE $\tilde s_{ij} \leftarrow
         \bigl\langle A_X(\tilde{\mathbf z}_{x,i}),
                     A_Y(\tilde{\mathbf z}_{y,j})\bigr\rangle / \tau$

  \STATE construct $\tilde q_i$ using Eq.~(6)      \hfill

  \STATE $\mathcal L$ by Eq.~(7)

  \STATE update parameters of $A_X$ and $A_Y$ with AdamW on $\mathcal L$
\end{algorithmic}
% \vspace{-1em}
\end{algorithm}
\vspace{-0.5em}

\subsection{Pipeline}

\textsc{Modest-Align} adjusts data matching by simulating input uncertainty and moderating model overconfidence in ambiguous positive samples, operating on the latent spaces \( \mathcal{U_X} \) and \( \mathcal{U_Y} \) derived from pre-trained unimodal encoders. 
1) The method initially employs unimodal encoders to encode V-L modalities into intermediate latent spaces. 2) Then, it utilizes enhanced features based on Fusemix in the \textsc{Modest-Align} latent spaces, incorporating \textbf{RP} to adjust data matching by simulating input uncertainty. 3) After training through VL-Adapters, \textbf{ES} is applied, aiming to smooth the model's prediction of output distributions, reduce overconfidence in positive pairs, and enhance the smoothness of predictions on uncertain samples. 4) Finally, the smoothed embeddings are used in contrastive learning training, facilitating the learning of two networks, \(A_X\) and \(A_Y\), as shown in \autoref{Figure_pipeline}.

\textsc{Modest-Align} utilizes the existing semantics encoded by unimodal encoders, reducing the reliance on extensive real paired data and simplifying computational requirements. 
It effectively mitigates the issue of model overconfidence, making the model more "modest" and robust, thereby optimizing cross-modal alignment, learning efficiency, and generalization capabilities.

\begin{table*}
\centering
\caption{The performance of SoTA methods and \textsc{Modest-Align} on different training datasets is assessed on the Flickr30K's 1K test set and MS-COCO's 5K test set, evaluating text-to-image and image-to-text retrieval accuracy using R@1 scores.}
\resizebox{\linewidth}{!}{
\label{experiment-table2}
\begin{tabular}{cccc|cc} 
\toprule
\multirow{2}{*}{\textbf{Training Dataset Size}}           & \multirow{2}{*}{\textbf{Method}} & \multicolumn{2}{c|}{Flickr30K (1K test set)}                                     & \multicolumn{2}{c}{MS-COCO (5K test set)}                                                                                                                                            \\ 
\cmidrule{3-6}
                                         &                                  & $\text{text} \rightarrow \text{image}$ & $\text{image} \rightarrow \text{text}$ & $\text{text} \rightarrow \text{image}$ & $\text{image} \rightarrow \text{text}$  \\ 
\midrule
{\cellcolor[rgb]{0.702,0.702,0.702}}400M & CLIP \cite{radford2021clip}             & 68.70                                  & 88.00                                  & \textcolor[rgb]{0.2,0.2,0.2}{37.80}                                                          & \textcolor[rgb]{0.2,0.2,0.2}{58.40}                                                     \\
{\cellcolor[rgb]{0.651,0.651,0.651}}4B   & LIT \cite{zhai2022lit}                  & 66.50                                  & 83.90                                  & \textcolor[rgb]{0.2,0.2,0.2}{43.60}                                                          & \textcolor[rgb]{0.2,0.2,0.2}{59.50}                                                     \\
{\cellcolor[rgb]{0.6,0.6,0.6}}5B         & 3T \cite{kossen2024three}               & 72.10                                  & 87.30                                  & 48.50                                                                                           & 64.10                                                                                     \\ 
\midrule
{\cellcolor[rgb]{0.949,0.949,0.949}}3M   & CLIP \cite{radford2021clip}             & 54.30                                  & 67.40                                  & 29.90                                                                                        & 36.20                                                                                   \\
{\cellcolor[rgb]{0.902,0.902,0.902}}3M   & Fusemix$_{(D,B)}$ \cite{vouitsis2024data} & 59.90                                  & 74.40                                  & 32.20                                                                                        & 42.30                                                                                   \\
{\cellcolor[rgb]{0.851,0.851,0.851}}5M (incl. 3.5M)   & Fusemix$_{(U,E)}$ \cite{vouitsis2024data} & 64.30                                  & 80.20                                  & 42.90                                                                                        & 59.10                                                                                   \\
{\cellcolor[rgb]{0.902,0.902,0.902}}3.5M & {\cellcolor{LightBlue}}\textsc{Modest-Align} (Ours)                  & {\cellcolor{LightBlue}}\textbf{65.72} \uaa{1.42}                         & {\cellcolor{LightBlue}}\textbf{81.60} \uaa{1.40}                         & \textcolor[rgb]{0.173,0.227,0.29}{{\cellcolor{LightBlue}}\textbf{45.86}\uaa{2.96}}                                            & \textcolor[rgb]{0.173,0.227,0.29}{{\cellcolor{LightBlue}}\textbf{62.86} \uaa{3.76}}                                      \\
\bottomrule
\end{tabular}
}
\vspace{-1em}
\end{table*}

\begin{figure*}[htbp]
\centering
% \label{logits_dataset}
\includegraphics[width=\textwidth]{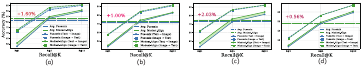}
\caption{Performance of \textsc{Modest-Align} and FuseMix \cite{vouitsis2024data} on Flickr30K (a, c) and MS-COCO (b, d) test sets.
Evaluated across training sets of varying scale (COCO for (a-b)), SBU for (c-d)), \textsc{Modest-Align} consistently demonstrates strong generalization and achieves SoTA performance on image retrieval benchmarks.
Red indicates average improvement.}
\label{experiment-table1}
\vspace{-1em}
\end{figure*}

\section{Experiments}

\textbf{Baselines and Metrics:} We conduct comparisons with several cross-modal alignment methods: Fusemix \cite{vouitsis2024data}, CLIP \cite{radford2021clip}, LIT \cite{zhai2022lit}, and 3T \cite{kossen2024three}, as shown in \autoref{experiment-table2} and \autoref{experiment-table1}. 
It's important to note that large-scale image-text datasets used by models like CLIP, LIT, and 3T, ranging from 400 million to 5 billion pairs, are mostly sourced from the internet and not publicly available, with high computational costs making them impractical for limited-resource scenarios. Therefore, Fusemix, offers a more applicable comparison with \textsc{Modest-Align} (\autoref{experiment-table1}).
% To assess the performance of cross-modal alignment for all considered methods, we use metrics such as R@1, R@5, and R@10. R@1 denotes Recall@1 for either text-to-image or image-to-text, R@5 indicates Recall@5, and R@10 stands for Recall@10, applicable to both text-to-image and image-to-text retrieval scenarios \cite{vouitsis2024data}.
We evaluate cross-modal alignment with Recall@K: R@1, R@5, and R@10—Recall at the top 1, 5, and 10 retrieved items for both text-to-image and image-to-text tasks \cite{vouitsis2024data}.
% we measure Top-1, Top-5, and Top-10 classification accuracy on the original validation set using models trained from scratch on these datasets.

\noindent\textbf{Experimental Setup:} 
% Since an important consideration of our method is to minimize computational requirements, we only use a single 24GB NVIDIA 3090 GPU for all of our experiments. 
% We can pre-compute the latents from pre-trained unimodal encoders so that the underlying encoders can be discarded thereafter. 
% Additionally, we can extract the latents for each modality one at a time to ensure that no more than one encoder must be loaded at once. 
To minimize computational demands, all our experiments are conducted on a single 24GB NVIDIA 3090 GPU. We pre-compute latents from pre-trained unimodal encoders, which are then discarded, extracting latents for each modality sequentially to avoid loading more than one encoder at a time. For consistency and fair comparison, we use the same unimodal encoders as Fusemix.
% We mainly consider Transformer-based \cite{vaswani2017attention} unimodal encoders, and extract low-dimensional latents from the penultimate layer of either the \texttt{[CLS]} token if it exists, or the mean-pooled token otherwise.  
% \textbf{Multimodal Latent Fusion.}
% We parameterize our fusion adapters as lightweight MLPs using an inverted bottleneck architecture following previous work \cite{lin2015far, tolstikhin2021mlp, bachmann2023scaling}. 
% Each MLP consists of residual blocks followed by a final projection layer of dimension 512 by default to embed each modality into a shared space. 
V-L adapters are parameterized as lightweight MLPs featuring an inverted bottleneck architecture, inspired by previous studies \cite{lin2015far, tolstikhin2021mlp, bachmann2023scaling}. Each MLP incorporates residual blocks and a default final projection layer with a dimension of 512, embedding each modality into a shared latent space.
For the image encoder, we consider DINOv2 \cite{oquab2023dinov2}, and for the text side, we select text encoder with demonstrably semantic latent spaces, specifically BGE \cite{xiao2023c}.
% We highlight that since our V-L adapters are operating on low-dimensional latents, the computational cost to train them is minimal, and despite training on a single GPU, we can use large batch sizes (up to $Batch=10$K on 3090 GPU), which has been shown to benefit contrastive learning \cite{wu2018unsupervised, tian2020contrastive, he2020momentum}. 
% The smoothing parameter \(\alpha\) is set to 0.1, and the Gaussian noise level \(\sigma\) is set to 0.01.
% \textcolor{red}{During experiments (see Appendix), we determined the optimal \(\sigma\) and \(\alpha\) values through grid search on the test set to ensure robust alignment. Results showed that \(\sigma\) (0.01) and \(\alpha\) (0.1) achieved the best performance.}
% \textcolor{red}{To systematically investigate the impact of batch size on performance, we plan to conduct ablation experiments with varying batch sizes (1k, 2k, 5k, 10k, and 15k). We will evaluate how these changes affect \textsc{Modest-Align}’s performance, especially on text-to-image and image retrieval metrics, to better understand the role of batch size in embedding smoothing. Our results indicate that a batch size of 10k offers optimal performance.}
Through grid search (see \textcolor{red}{Appendix}), we found optimal values of \(\sigma=0.01\) and \(\alpha=0.1\), and our ablation experiments show that a batch size of 10k provides the best performance for \textsc{Modest-Align}.
% Finally, we note that in all of our experiments, unless otherwise stated, we use $\mathcal{L}_\text{sym}^\text{FuseMix}$ as our sole objective for multimodal fusion. More details on the MLP architecture and hyperparameters can be found in Appendix \ref{sec:appendix-arch} and \ref{sec:appendix-impl}.

\noindent\textbf{Training and Test Datasets:} 
To evaluate the effectiveness of the \textsc{Modest-Align} for the task of modality alignment, we conducted extensive comparative experiments against SoTA methods across various datasets. following previous works \cite{chen2020uniter, li2021albef, li2022blip, li2023blip2},  
These training datasets include COCO \cite{lin2014mscoco}, Visual Genome (VG) \cite{krishna2017visualgenome}, SBU \cite{ordonez2011sbucaptions}, and Conceptual Captions 3M (CC3M) \cite{sharma2018conceptualcaptions}. \autoref{tab:Distribution of logits} provides detailed information about these four datasets. 
It is noteworthy that the original CC3M dataset, consisting of images stored as internet URLs, currently has only 1.5 million data pairs available.
% Utilizing a single NVIDIA 3090 GPU for training, \textsc{Modest-Align} demonstrated SoTA performance across datasets of varying sizes. The specific results and analyses will be discussed in the subsequent sections.
We tested \textsc{Modest-Align} on the Flickr \cite{young2014image} and MS-COCO datasets \cite{lin2014mscoco} to benchmark performance in image-text retrieval and assess generalization across scales.

\begin{figure}[t]
\centering
% \label{logits_dataset}
\hspace{-6mm} 
\includegraphics[width=0.51\textwidth]{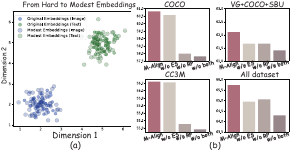}
\caption{
(a) In the original distribution, clusters of image-text pairs are tightly packed, showing high model confidence. After implementing \textsc{Modest-Align}, the embeddings become more dispersed, indicating reduced confidence in individual labels and a more robust, softer probability distribution that enhances model generalization. 
(b) The ablation study of two techniques (RP and ES) in Recall@1.
} 
\label{Figure_modest}
\vspace{-1.5em}
% \hspace{-5em}
\end{figure}

\subsection{Results and Analysis}

\textsc{Modest-Align} demonstrates a significant advantage in achieving high performance with substantially lower training costs. 
% Compared to models of similar size, \textsc{Modest-Align} surpasses both CLIP and Fusemix, achieving a R@1 improvement of 20.98\% in text-to-image retrieval and 21.08\% in image-to-text retrieval over CLIP, and improvements of 9.7\% over Fusemix in both tasks, as shown in \autoref{experiment-table2}. 
Compared to training datasets of similar size, \textsc{Modest-Align} outperforms both CLIP and Fusemix, achieving an R@1 improvement of 11.42\% in text-to-image retrieval and 14.2\% in image-to-text retrieval over CLIP (\autoref{experiment-table2}). 
Additionally, it surpasses Fusemix with improvements of 5.82\% and 7.2\% in both tasks, respectively, as detailed in \autoref{experiment-table2}.
Furthermore, despite using significantly smaller training data (3.5M pairs), \textsc{Modest-Align} performs competitively against much larger datasets used by CLIP (400M) and LIT (4B), trailing by only 0.78\% in text-to-image and 2.30\% in Flickr image-to-text retrieval against LIT. This demonstrates the effectiveness and efficiency of \textsc{Modest-Align}. 
% We anticipate that with further increases in training data, \textsc{Modest-Align} has the potential to surpass these SoTA methods, further underscoring its usability.
% \textcolor{red}{Additionally, we tested on the MS-COCO dataset, demonstrating its superior performance across various datasets.  It has outperformed state-of-the-art methods on public datasets, notably improving the R@1 score in retrieval tasks. With only 3.5M data, \textsc{Modest-Align} exceeds CLIP’s performance on the MS-COCO dataset and does so with 100 times less data and over 600 times less training time than CLIP, which requires 3000 GPU days and 400M training data.
% }
We tested \textsc{Modest-Align} on the MS-COCO, demonstrating its superior performance across various datasets, particularly improving the R@1 score. With 3.5M data, \textsc{Modest-Align} outperforms CLIP on MS-COCO, requiring 100 times less data and over 600 times less training time, compared to CLIP’s 3000 GPU days and 400M data (\autoref{Figure_days}). Classification results of \textsc{Modest-Align} versus CLIP are reported in the Appendix.

As shown in \autoref{experiment-table1}, \textsc{Modest-Align} consistently outperforms Fusemix across various dataset sizes and settings in image-text retrieval tasks. 
On the COCO ($560K$ pairs), \textsc{Modest-Align} achieves a significant improvement in text-to-image retrieval, with a 3\% higher R@1 score compared to Fusemix. 
% {($3\% = 60.8\% -57.8\%$)}. 
% This advantage is even more pronounced on larger datasets. 
On SBU ($840K$ pairs), \textsc{Modest-Align} surpasses Fusemix by over 4.48\% in R@1 for text-to-image tasks.
% image-to-text tasks. 
% Furthermore, on the four dataset training datasets configuration (VG+COCO+SBU+CC3M, 3.5M samples), \textsc{Modest-Align} delivers more than a 1.5\% improvement in R@1 for text-to-image retrieval and a substantial 2.5\% increase in image-to-text retrieval. 
% \textsc{Modest-Align} improves R@1 by over 0.7\% on a combined training of VG, COCO, and SBU ($2M$ pairs).
Furthermore, on a combined training configuration of four datasets (VG+COCO+SBU+CC3M, totaling $3.5M$ pairs), \textsc{Modest-Align} achieves a significant improvement of over 1.44\% in R@1 for text-to-image retrieval.
Besides, we tested on the MS-COCO, demonstrating its superior performance across various datasets.
These results highlight \textsc{Modest-Align}'s robust generalization capabilities and superior performance over the SoTA method.

Using CLIP-ViT/B-32 \cite{radford2021clip}, we assessed image-text quality and found that most pairs are only weakly aligned: 93.5 \% of CC3M, 91.9 \% of COCO, 94.4 \% of SBU, and 99.1 \% of VG have cosine scores < 35 (\autoref{tab:Distribution of logits}, \autoref{Figure_datasets}).
COCO’s comparatively higher alignment explains its stronger downstream results (\autoref{Figure_datasets}).
Such low-quality, ambiguous pairs drive over-confidence in standard contrastive training and mislead inference, motivating \textsc{Modest-Align}.

\subsection{Ablation Study}

\begin{figure}[t]
\centering
% \label{logits_dataset}
% \hspace{-6mm} 
\includegraphics[width=0.50\textwidth]{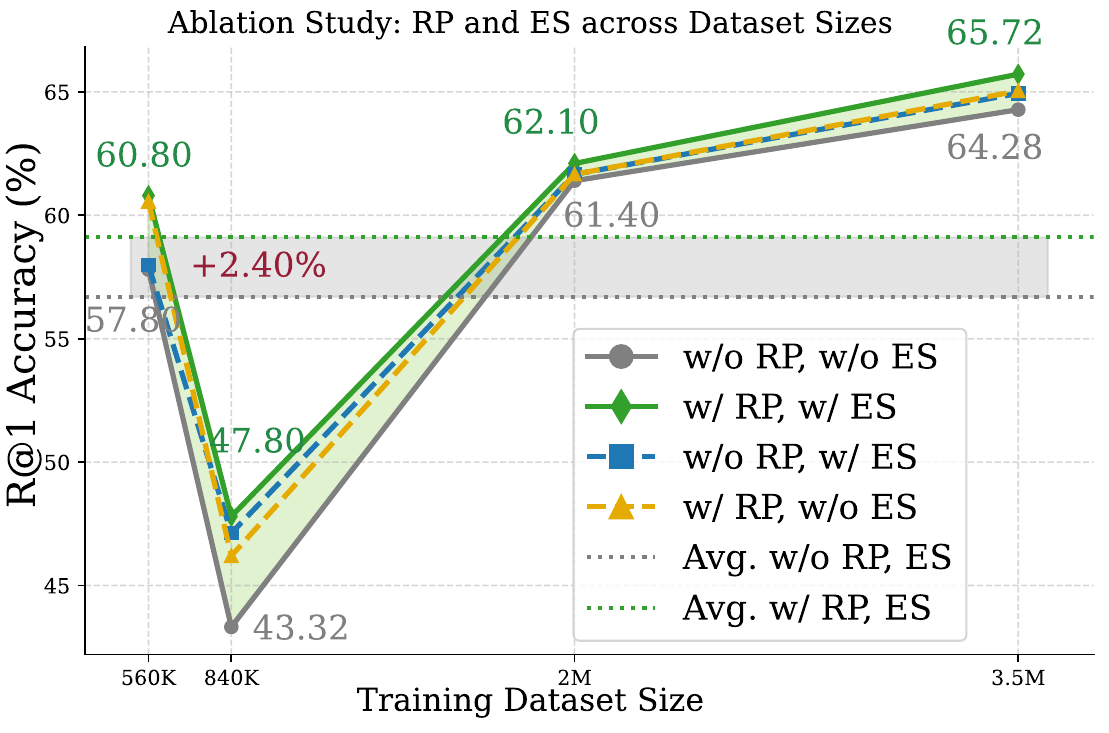}
\caption{
Ablation experiments of different techniques, including RP and ES, are conducted across various datasets, measuring text-to-image retrieval performance (R@1).
} 
\label{experiment-table}
\vspace{-1em}
% \hspace{-5em}
\end{figure}

\textbf{Effect of RP:} 
To validate the effectiveness of RP, we conducted ablation experiments for RP.
As shown in \autoref{experiment-table}, adding RP consistently improves performance across all datasets. 
% On the VG+COCO+SBU+CC3M, the R@1 score increases from 64.28\% to 65.72, a 1.44\% improvement with PR. 
% On the VG+COCO+SBU+CC3M datasets, adding Random Perturbation (RP) in two settings—with and without Embedding Smoothing (ES)—increases the R@1 score from 64.94 to 65.72, a 0.78\% improvement, and from 64.28 to 65.04, a 0.76\% improvement.
On the COCO, adding RP in two settings—with and without ES—results in an R@1 score increase from 57.98\% to 60.8\%, a 2.82\% improvement, and from 57.8\% to 60.56\%, a 2.76\% improvement. Similarly, this positive trend in gains from RP is observable across other datasets as well, as shown in \autoref{Figure_modest} (b).
% Similarly, on the VG+COCO+SUB dataset, 
% the score rises by 1.1\%, from 61.4\% to 62.1\%. 
% The effect is more pronounced on larger datasets, with a 10.3\% improvement on SUB dataset, where the score jumps from 43.32\% to 47.8\%. 
% These results demonstrate that RP significantly enhances model performance, particularly for specific datasets.
% In addition, through the ablation experiment, it is found that adding RP technology while using Embedding Smoothing(ES) has a gain effect on the final task.
These results demonstrate that RP significantly enhances model performance regardless of whether ES techniques are used.

\noindent\textbf{Effect of ES:} 
% ES consistently enhances text-to-image retrieval performance, especially on larger datasets. 
% As shown in \autoref{experiment-table}, ES improves R@1 score on the COCO dataset by 4.8\% and , 
% and on the VG+COCO+SBU+CC3M dataset, it leads to a substantial 1.18\% increase. 
As shown in \autoref{experiment-table}, ES improves the R@1 on the SBU by 3.8\% without RP and by 1.6\% with RP. Similar gains from ES are observed across other datasets, regardless of the RP, as shown in \autoref{Figure_modest}.
ES enhances performance on complex, large-scale datasets by improving generalization across diverse image-text pairs. 
% As a regularization strategy in contrastive learning, it smooths target distributions to mitigate overfitting, thereby boosting model robustness and accuracy on validation datasets.
As shown in \autoref{tab:dynamic value ES}, Default ES outperforms dynamic $\alpha$ ($\alpha$ is dynamically decayed over training epochs from 0.1 to 0.01 to progressively reduce supervision rigidity), achieving a 0.34\% improvement in R@1 for text-to-image retrieval and 0.8\% for image-to-text on COCO. 
These gains are consistent across datasets, with default ES proving more effective in optimizing retrieval tasks and handling complex data. Both smoothing methods outperform models without ES, highlighting its effectiveness.

\newcommand{\cmark}{\ding{51}} % check mark
\newcommand{\xmark}{\ding{55}} % cross mark

\begin{table}[t]
\centering
\caption{
% Ablation Study of Different ES Types in Various Datasets (R@1). This study compares the performance of different types of Embedding Smoothing (ES) across several datasets. 
% We use \textbf{\xmark} to denote without ES, Dynamic for dynamic $\alpha$ in ES, and \textbf{\cmark} for default $\alpha$ in ES.
Ablation study on ES variants across datasets (R@1).
\xmark, denotes training without ES, Dynamic uses adaptive $\alpha$, and \cmark, indicates default $\alpha$.
}
\label{tab:dynamic value ES}
\small
 \resizebox{\linewidth}{!}{
\begin{tabular}{cccc|ccc} 
\toprule
   \multirow{3}{*}{Dataset Size}              & \multicolumn{6}{c}{Different Type of ES}                                                                                                                \\ 
\cmidrule{2-7}
                 & \multicolumn{3}{c|}{$\text{text} \rightarrow \text{image}$}                & \multicolumn{3}{c}{$\text{image} \rightarrow \text{text}$}                 \\
          & \textbf{\xmark} & Dynamic & \textbf{\cmark}                                         & \textbf{\xmark} & Dynamic & \textbf{\cmark}                                         \\ 
\midrule
$560K$             & 57.80      & 60.22      & {\cellcolor[rgb]{0.969,0.988,1}}\textbf{60.56} & 71.60      & 72.70       & {\cellcolor[rgb]{0.969,0.988,1}}\textbf{73.50}  \\
$820K$               & 51.66     & 52.50       & {\cellcolor[rgb]{0.969,0.988,1}}\textbf{53.80}  & 66.90      & 68.10       & {\cellcolor[rgb]{0.969,0.988,1}}\textbf{69.20}  \\
$840K$             & 43.32     & 45.46      & {\cellcolor[rgb]{0.969,0.988,1}}\textbf{47.12} & 61.60      & 63.60       & {\cellcolor[rgb]{0.969,0.988,1}}\textbf{62.80}    \\
$2M$      & 61.40      & 61.32      & {\cellcolor[rgb]{0.969,0.988,1}}\textbf{61.66} & 77.20      & 77.90       & {\cellcolor[rgb]{0.969,0.988,1}}\textbf{78.40}  \\
$3.5M$ & 64.28     & 65.14      & {\cellcolor[rgb]{0.969,0.988,1}}{64.94}          & 81.20      & 81.50       & {\cellcolor[rgb]{0.969,0.988,1}}\textbf{82.50}  \\
\bottomrule
\end{tabular}
}
\vspace{-1em}
\end{table}

% \usepackage{colortbl}

 % \in (0, 1) \) is the 
% \textbf{Effect of Smoothing Parameter $\alpha$ in ES:} 
% As shown in \autoref{tab:dynamic value ES}, default parameter Embedding Smoothing (ES) outperforms Dynamic \(\alpha\) ES (\textbf{◯}, as training progresses, \(\alpha\) gradually decreases) in the majority of cases.
% the ablation study in \autoref{tab:dynamic value ES} highlights the superior adaptability and effectiveness of default $\alpha$.
% As shown in \autoref{tab:dynamic value ES}, default parameter ES consistently outperforms Dynamic \(\alpha\) of ES (with \(\alpha\) decreasing over training), demonstrating the default \(\alpha\)'s effectiveness.
% On the COCO dataset, default ES achieves a marginal but notable 0.34\% improvement in R@1 scores for text-to-image retrieval, while the improvement in image-to-text retrieval is more substantial, with a 0.8\% increase. 
% These results are consistent across other datasets, where default $\alpha$ ES consistently delivers performance enhancements. 
% While dynamic ES offers certain benefits, default ES is more effective in optimizing retrieval tasks, particularly by adapting to diverse datasets and handling complex data more efficiently.
% Both dynamic and default smoothing parameters significantly outperform models without ES, underscoring ES's effectiveness.

\noindent\textbf{Impact of Dataset Quality:} 
Human-annotated COCO (\(560K\)) outperforms larger SBU ((\(840K\)) in retrieval, achieving 57.80\% vs. 43.32\% R@1, regardless of the method used (\autoref{tab:performance_comparison}).
This shows that dataset curation is more impactful than simply increasing size. \textsc{Modest-Align}’s ability to adjust image-text matching is valuable, especially since large-scale curation is impractical. 
% As shown in \autoref{Figure_modest} (a), \textsc{Modest-Align} shifts embeddings from “hard” to “modest,” making them smoother and more dispersed, which reduces confidence in individual samples and enhances model generalization.
As shown in \autoref{Figure_modest} (a), \textsc{Modest-Align} softens hard embeddings into smoother, more dispersed ones, lowering overconfidence and improving generalization.

\begin{table}
\centering
\setlength{\extrarowheight}{0pt}
\addtolength{\extrarowheight}{\aboverulesep}
\addtolength{\extrarowheight}{\belowrulesep}
\setlength{\aboverulesep}{0pt}
\setlength{\belowrulesep}{0pt}
\caption{
% Analysis of the Impact of Dataset Quality and Size with \textcolor{red}{FuseMix} on Flickr30K. This analysis compares the effects of varying dataset quality and size on the \textsc{Modest-Align}'s performance. The smaller datasets of higher quality outperform larger but lower-quality datasets, underscoring the importance of data quality. 
Impact of dataset quality and size on Flickr30K retrieval.
We compare {Fusemix} (light red) and {\textsc{Modest-Align}} (light blue) under varying data conditions. Smaller, high-quality datasets outperform larger but noisier ones, highlighting the importance of data quality.
}
\label{tab:performance_comparison}
\resizebox{\linewidth}{!}{
\begin{tabular}{ccccc|ccc} 
\toprule
\multirow{2}{*}{Size}                    & \multirow{2}{*}{Dataset} & \multicolumn{3}{c|}{$\text{text} \rightarrow \text{image}$} & \multicolumn{3}{c}{$\text{image} \rightarrow \text{text}$}  \\ 
\cmidrule{3-8}
                                         &                          & R@1           & R@5           & R@10                        & R@1           & R@5           & R@10                        \\ 
\midrule
\rowcolor[rgb]{0.992,0.973,0.973} $840K$ & SBU                      & \uline{43.32} & \uline{72.48} & \uline{81.36}               & \uline{61.60} & \uline{86.30} & \uline{92.10}               \\
\rowcolor[rgb]{0.992,0.973,0.973} $560K$ & COCO                     & 57.80         & 83.38         & 89.54                       & 71.60         & 91.10         & 95.00                       \\ 
\midrule
\rowcolor[rgb]{0.969,0.988,1} $840K$     & SBU                      & \uline{47.80} & \uline{75.94} & \uline{84.18}               & \uline{62.10} & \uline{87.00} & \uline{92.30}               \\
\rowcolor[rgb]{0.969,0.988,1} $560K$     & COCO                     & 60.80         & 84.82         & 90.82                       & 72.60         & 93.30         & 95.70                       \\
\bottomrule
\end{tabular}
}
% \vspace{-1em}
\end{table}

\subsection{Discussion}
% ES, inspired by label smoothing, is specifically designed for cross-modal alignment tasks, addressing overconfidence in image-text pairs. Unlike label smoothing, which is applied to unimodal classification tasks to adjust category probabilities, ES operates within the embedding space, smoothing the similarity distribution between positive and negative pairs. This enhances robustness to weakly correlated samples, improving generalization in cross-modal alignment.
ES, inspired by label smoothing, adapts its core idea to the embedding space for cross-modal alignment. By smoothing the similarity distribution between positive and negative pairs, ES mitigates overconfidence and improves robustness to weakly correlated samples.
RP’s uniqueness lies in its primary goal to simulate uncertainty in data pairs for cross-modal alignment tasks, enhancing model robustness. 
RP is designed for the embedding space, helping the model learn more “modestly” on weakly correlated datasets. 
% This approach is particularly suited for multimodal tasks, rather than traditional unimodal tasks.
\textsc{Modest-Align} is the first to unify them under a cohesive objective tailored for cross-modal uncertainty, enabling robust contrastive learning over noisy or weakly aligned data. 
% The combination forms a principled and novel loss formulation optimized for this unique alignment scenario.
This combination yields a principled loss tailored for robust alignment under uncertainty.

\section{Conclusion}

In this work, we propose \textsc{Modest-Align}, a data- and compute-efficient cross-modal alignment method that mitigates overconfidence, strengthens latent pairwise associations, and leverages pretrained unimodal encoders for guidance.
% Notably, \textsc{Modest-Align} excels on datasets with lower data quality (image-text match level) and enhances performance on datasets with higher match levels. 
Validated across multiple datasets, \textsc{Modest-Align} has consistently demonstrated its robust capability to align V-L models better.
% In the future, we could consider developing \textsc{Modest-Align} to assess data quality, allowing for dynamic adjustment of model confidence by applying stricter constraints on lower quality data and less stringent constraints on higher quality data to dynamically enhance or reduce model overconfidence.

\newpage

\subsubsection*{Limitations}
% Although \textsc{Modest-Align} achieved performance improvements across all datasets and their combinations, it is notable that the gains are more significant for datasets of poorer quality, while the enhancements are more modest for relatively better datasets. 
% We are unable to test \textsc{Modest-Align} on larger datasets like the 400M pairs used in CLIP, as it is not publicly available. Consequently, it is difficult to ascertain \textsc{Modest-Align}'s performance benefits for extremely large-scale datasets. 
% Besides, it remains unclear how much \textsc{Modest-Align} benefits datasets of very high quality, where image-text pairs are perfectly matched.

% \textsc{Modest-Align}, trained with a single GPU, improved performance and alignment across all datasets and their combinations under limited training and data resources. We observed that the gains were more significant for lower-quality datasets, while the improvements were more modest for higher-quality datasets. 
Trained on a single GPU, \textsc{Modest-Align} consistently improved alignment across all datasets, with larger gains on lower-quality data.
One limitation of \textsc{Modest-Align} is that its performance on extremely large-scale datasets remains unclear, as training such datasets is impractical under limited GPU resources. Additionally, it is uncertain how much \textsc{Modest-Align} would benefit datasets with very high-quality, perfectly matched image-text pairs, because such data is rare.

For future developments, \textsc{Modest-Align} could incorporate data quality assessments to dynamically adjust model confidence, applying stricter constraints on lower quality data and more lenient ones on higher quality data to effectively manage model overconfidence.
\bibliography{custom}

\begin{thebibliography}{51}
\providecommand{\natexlab}[1]{#1}

\bibitem[{Alayrac et~al.(2022)Alayrac, Donahue, Luc, Miech, Barr, Hasson, Lenc, Mensch, Millican, Reynolds, Ring, Rutherford, Cabi, Han, Gong, Samangooei, Monteiro, Menick, Borgeaud, Brock, Nematzadeh, Sharifzadeh, Bi\'{n}kowski, Barreira, Vinyals, Zisserman, and Simonyan}]{alayrac2022flamingo}
Jean-Baptiste Alayrac, Jeff Donahue, Pauline Luc, Antoine Miech, Iain Barr, Yana Hasson, Karel Lenc, Arthur Mensch, Katherine Millican, Malcolm Reynolds, Roman Ring, Eliza Rutherford, Serkan Cabi, Tengda Han, Zhitao Gong, Sina Samangooei, Marianne Monteiro, Jacob~L Menick, Sebastian Borgeaud, Andy Brock, Aida Nematzadeh, Sahand Sharifzadeh, Miko\l~aj Bi\'{n}kowski, Ricardo Barreira, Oriol Vinyals, Andrew Zisserman, and Kar\'{e}n Simonyan. 2022.
\newblock {Flamingo: a Visual Language Model for Few-Shot Learning}.
\newblock In \emph{Advances in Neural Information Processing Systems}, volume~35, pages 23716--23736.

\bibitem[{Arandjelovic and Zisserman(2017)}]{arandjelovic2017looklistenlearn}
Relja Arandjelovic and Andrew Zisserman. 2017.
\newblock Look, listen and learn.
\newblock In \emph{Proceedings of the IEEE International Conference on Computer Vision}.

\bibitem[{Bachmann et~al.(2023)Bachmann, Anagnostidis, and Hofmann}]{bachmann2023scaling}
Gregor Bachmann, Sotiris Anagnostidis, and Thomas Hofmann. 2023.
\newblock {Scaling MLPs: A Tale of Inductive Bias}.
\newblock \emph{arXiv:2306.13575}.

\bibitem[{Bao et~al.(2022)Bao, Wang, Dong, Liu, Mohammed, Aggarwal, Som, Piao, and Wei}]{bao2022vlmo}
Hangbo Bao, Wenhui Wang, Li~Dong, Qiang Liu, Owais~Khan Mohammed, Kriti Aggarwal, Subhojit Som, Songhao Piao, and Furu Wei. 2022.
\newblock {VLMo: Unified Vision-Language Pre-Training with Mixture-of-Modality-Experts}.
\newblock In \emph{Advances in Neural Information Processing Systems}, volume~35, pages 32897--32912.

\bibitem[{Chen et~al.(2020)Chen, Li, Yu, El~Kholy, Ahmed, Gan, Cheng, and Liu}]{chen2020uniter}
Yen-Chun Chen, Linjie Li, Licheng Yu, Ahmed El~Kholy, Faisal Ahmed, Zhe Gan, Yu~Cheng, and Jingjing Liu. 2020.
\newblock {UNITER: UNiversal Image-TExt Representation Learning}.
\newblock In \emph{Computer Vision -- ECCV 2020}, pages 104--120.

\bibitem[{Girdhar et~al.(2023)Girdhar, El-Nouby, Liu, Singh, Alwala, Joulin, and Misra}]{girdhar2023imagebind}
Rohit Girdhar, Alaaeldin El-Nouby, Zhuang Liu, Mannat Singh, Kalyan~Vasudev Alwala, Armand Joulin, and Ishan Misra. 2023.
\newblock {ImageBind: One Embedding Space To Bind Them All}.
\newblock In \emph{Proceedings of the IEEE/CVF Conference on Computer Vision and Pattern Recognition}, pages 15180--15190.

\bibitem[{Girdhar et~al.(2022)Girdhar, Singh, Ravi, van~der Maaten, Joulin, and Misra}]{girdhar2022omnivore}
Rohit Girdhar, Mannat Singh, Nikhila Ravi, Laurens van~der Maaten, Armand Joulin, and Ishan Misra. 2022.
\newblock Omnivore: A single model for many visual modalities.
\newblock In \emph{Proceedings of the IEEE/CVF Conference on Computer Vision and Pattern Recognition}, pages 16102--16112.

\bibitem[{Jia et~al.(2021)Jia, Yang, Xia, Chen, Parekh, Pham, Le, Sung, Li, and Duerig}]{jia2021align}
Chao Jia, Yinfei Yang, Ye~Xia, Yi-Ting Chen, Zarana Parekh, Hieu Pham, Quoc Le, Yun-Hsuan Sung, Zhen Li, and Tom Duerig. 2021.
\newblock Scaling up visual and vision-language representation learning with noisy text supervision.
\newblock In \emph{Proceedings of the 38th International Conference on Machine Learning}, volume 139, pages 4904--4916. PMLR.

\bibitem[{Kossen et~al.(2024)Kossen, Collier, Mustafa, Wang, Zhai, Beyer, Steiner, Berent, Jenatton, and Kokiopoulou}]{kossen2024three}
Jannik Kossen, Mark Collier, Basil Mustafa, Xiao Wang, Xiaohua Zhai, Lucas Beyer, Andreas Steiner, Jesse Berent, Rodolphe Jenatton, and Effrosyni Kokiopoulou. 2024.
\newblock Three towers: Flexible contrastive learning with pretrained image models.
\newblock \emph{Advances in Neural Information Processing Systems}, 36.

\bibitem[{Kossen et~al.(2023)Kossen, Collier, Mustafa, Wang, Zhai, Beyer, Steiner, Berent, Jenatton, and Kokiopoulou}]{kossen2023three}
Jannik Kossen, Mark Collier, Basil Mustafa, Xiao Wang, Xiaohua Zhai, Lucas Beyer, Andreas Steiner, Jesse Berent, Rodolphe Jenatton, and Efi Kokiopoulou. 2023.
\newblock Three towers: Flexible contrastive learning with pretrained image models.
\newblock \emph{arXiv:2305.16999}.

\bibitem[{Krishna et~al.(2017{\natexlab{a}})Krishna, Zhu, Groth, Johnson, Hata, Kravitz, Chen, Kalantidis, Li, Shamma, Bernstein, and Li}]{krishna2017visualgenome}
Ranjay Krishna, Yuke Zhu, Oliver Groth, Justin Johnson, Kenji Hata, Joshua Kravitz, Stephanie Chen, Yannis Kalantidis, Li-Jia Li, David~A Shamma, Michael~S. Bernstein, and Fei-Fei Li. 2017{\natexlab{a}}.
\newblock {Visual genome: Connecting language and vision using crowdsourced dense image annotations}.
\newblock \emph{International Journal of Computer Vision}, 123:32--73.

\bibitem[{Krishna et~al.(2017{\natexlab{b}})Krishna, Zhu, Groth, Johnson, Hata, Kravitz, Chen, Kalantidis, Li, Shamma et~al.}]{krishna2017visual}
Ranjay Krishna, Yuke Zhu, Oliver Groth, Justin Johnson, Kenji Hata, Joshua Kravitz, Stephanie Chen, Yannis Kalantidis, Li-Jia Li, David~A Shamma, et~al. 2017{\natexlab{b}}.
\newblock Visual genome: Connecting language and vision using crowdsourced dense image annotations.
\newblock \emph{International journal of computer vision}, 123:32--73.

\bibitem[{Li et~al.(2023)Li, Li, Savarese, and Hoi}]{li2023blip2}
Junnan Li, Dongxu Li, Silvio Savarese, and Steven Hoi. 2023.
\newblock {BLIP}-2: Bootstrapping language-image pre-training with frozen image encoders and large language models.
\newblock In \emph{Proceedings of the 40th International Conference on Machine Learning}. PMLR.

\bibitem[{Li et~al.(2022)Li, Li, Xiong, and Hoi}]{li2022blip}
Junnan Li, Dongxu Li, Caiming Xiong, and Steven Hoi. 2022.
\newblock {BLIP}: Bootstrapping language-image pre-training for unified vision-language understanding and generation.
\newblock In \emph{Proceedings of the 39th International Conference on Machine Learning}, volume 162, pages 12888--12900. PMLR.

\bibitem[{Li et~al.(2021)Li, Selvaraju, Gotmare, Joty, Xiong, and Hoi}]{li2021albef}
Junnan Li, Ramprasaath Selvaraju, Akhilesh Gotmare, Shafiq Joty, Caiming Xiong, and Steven Chu~Hong Hoi. 2021.
\newblock Align before fuse: Vision and language representation learning with momentum distillation.
\newblock In \emph{Advances in Neural Information Processing Systems}, volume~34, pages 9694--9705.

\bibitem[{Li et~al.(2020)Li, Yin, Li, Zhang, Hu, Zhang, Wang, Hu, Dong, Wei, Choi, and Gao}]{li2020oscar}
Xiujun Li, Xi~Yin, Chunyuan Li, Pengchuan Zhang, Xiaowei Hu, Lei Zhang, Lijuan Wang, Houdong Hu, Li~Dong, Furu Wei, Yejin Choi, and Jianfeng Gao. 2020.
\newblock {Oscar: Object-Semantics Aligned Pre-training for Vision-Language Tasks}.
\newblock In \emph{Computer Vision -- ECCV 2020}, pages 121--137.

\bibitem[{Likhosherstov et~al.(2023)Likhosherstov, Arnab, Choromanski, Lucic, Tay, and Dehghani}]{likhosherstov2023polyvit}
Valerii Likhosherstov, Anurag Arnab, Krzysztof~Marcin Choromanski, Mario Lucic, Yi~Tay, and Mostafa Dehghani. 2023.
\newblock {PolyViT: Co-training Vision Transformers on Images, Videos and Audio}.
\newblock \emph{Transactions on Machine Learning Research}.

\bibitem[{Lin et~al.(2014{\natexlab{a}})Lin, Maire, Belongie, Hays, Perona, Ramanan, Doll{\'a}r, and Zitnick}]{lin2014mscoco}
Tsung-Yi Lin, Michael Maire, Serge Belongie, James Hays, Pietro Perona, Deva Ramanan, Piotr Doll{\'a}r, and C.~Lawrence Zitnick. 2014{\natexlab{a}}.
\newblock {Microsoft COCO: Common Objects in Context}.
\newblock In \emph{Computer Vision -- ECCV 2014}, pages 740--755. Springer International Publishing.

\bibitem[{Lin et~al.(2014{\natexlab{b}})Lin, Maire, Belongie, Hays, Perona, Ramanan, Doll{\'a}r, and Zitnick}]{lin2014microsoft}
Tsung-Yi Lin, Michael Maire, Serge Belongie, James Hays, Pietro Perona, Deva Ramanan, Piotr Doll{\'a}r, and C~Lawrence Zitnick. 2014{\natexlab{b}}.
\newblock Microsoft coco: Common objects in context.
\newblock In \emph{Computer Vision--ECCV 2014: 13th European Conference, Zurich, Switzerland, September 6-12, 2014, Proceedings, Part V 13}, pages 740--755. Springer.

\bibitem[{Lin et~al.(2015)Lin, Memisevic, and Konda}]{lin2015far}
Zhouhan Lin, Roland Memisevic, and Kishore Konda. 2015.
\newblock How far can we go without convolution: Improving fully-connected networks.
\newblock \emph{arXiv:1511.02580}.

\bibitem[{Liu et~al.(2023{\natexlab{a}})Liu, Hao, Lin, Pan, Yang, Feng, Wang, Li, Jin, Zhao et~al.}]{liu2023deep}
Jiaxiang Liu, Jin Hao, Hangzheng Lin, Wei Pan, Jianfei Yang, Yang Feng, Gaoang Wang, Jin Li, Zuolin Jin, Zhihe Zhao, et~al. 2023{\natexlab{a}}.
\newblock Deep learning-enabled 3d multimodal fusion of cone-beam ct and intraoral mesh scans for clinically applicable tooth-bone reconstruction.
\newblock \emph{Patterns}, 4(9).

\bibitem[{Liu et~al.(2025)Liu, Hu, Du, Zhang, Zhou, and Liu}]{liu2025kpl}
Jiaxiang Liu, Tianxiang Hu, Jiawei Du, Ruiyuan Zhang, Joey~Tianyi Zhou, and Zuozhu Liu. 2025.
\newblock Kpl: Training-free medical knowledge mining of vision-language models.
\newblock \emph{arXiv preprint arXiv:2501.11231}.

\bibitem[{Liu et~al.(2024{\natexlab{a}})Liu, Hu, Xiong, Du, Feng, Wu, Zhou, and Liu}]{liu2024vpl}
Jiaxiang Liu, Tianxiang Hu, Huimin Xiong, Jiawei Du, Yang Feng, Jian Wu, Joey Zhou, and Zuozhu Liu. 2024{\natexlab{a}}.
\newblock Vpl: Visual proxy learning framework for zero-shot medical image diagnosis.
\newblock In \emph{Findings of the Association for Computational Linguistics: EMNLP 2024}, pages 9978--9992.

\bibitem[{Liu et~al.(2023{\natexlab{b}})Liu, Hu, Zhang, Feng, Hao, Lv, and Liu}]{liu2023parameter}
Jiaxiang Liu, Tianxiang Hu, Yan Zhang, Yang Feng, Jin Hao, Junhui Lv, and Zuozhu Liu. 2023{\natexlab{b}}.
\newblock Parameter-efficient transfer learning for medical visual question answering.
\newblock \emph{IEEE Transactions on Emerging Topics in Computational Intelligence}.

\bibitem[{Liu et~al.(2023{\natexlab{c}})Liu, Hu, Zhang, Gai, Feng, and Liu}]{liu2023chatgpt}
Jiaxiang Liu, Tianxiang Hu, Yan Zhang, Xiaotang Gai, Yang Feng, and Zuozhu Liu. 2023{\natexlab{c}}.
\newblock A chatgpt aided explainable framework for zero-shot medical image diagnosis.
\newblock \emph{arXiv preprint arXiv:2307.01981}.

\bibitem[{Liu et~al.(2024{\natexlab{b}})Liu, Wang, Du, Zhou, and Liu}]{liu2024medcot}
Jiaxiang Liu, Yuan Wang, Jiawei Du, Joey Zhou, and Zuozhu Liu. 2024{\natexlab{b}}.
\newblock Medcot: Medical chain of thought via hierarchical expert.
\newblock In \emph{Proceedings of the 2024 Conference on Empirical Methods in Natural Language Processing}, pages 17371--17389.

\bibitem[{Loshchilov and Hutter(2016)}]{loshchilov2016sgdr}
Ilya Loshchilov and Frank Hutter. 2016.
\newblock Sgdr: Stochastic gradient descent with warm restarts.
\newblock \emph{arXiv preprint arXiv:1608.03983}.

\bibitem[{Loshchilov and Hutter(2018)}]{loshchilov2018fixing}
Ilya Loshchilov and Frank Hutter. 2018.
\newblock Fixing weight decay regularization in adam.

\bibitem[{Lu et~al.(2019)Lu, Batra, Parikh, and Lee}]{lu2019vilbert}
Jiasen Lu, Dhruv Batra, Devi Parikh, and Stefan Lee. 2019.
\newblock {ViLBERT: Pretraining Task-Agnostic Visiolinguistic Representations for Vision-and-Language Tasks}.
\newblock In \emph{Advances in Neural Information Processing Systems}, volume~32.

\bibitem[{Oquab et~al.(2023)Oquab, Darcet, Moutakanni, Vo, Szafraniec, Khalidov, Fernandez, Haziza, Massa, El-Nouby, Assran, Ballas, Galuba, Huang, Shang-Wen, Misra, Rabbat, Sharma, Synnaeve, Xu, Jegou, Mairal, Labatut, Joulin, and Bojanowski}]{oquab2023dinov2}
Maxime Oquab, Timoth{\'e}e Darcet, Th{\'e}o Moutakanni, Huy Vo, Marc Szafraniec, Vasil Khalidov, Pierre Fernandez, Daniel Haziza, Francisco Massa, Alaaeldin El-Nouby, Mahmoud Assran, Nicolas Ballas, Russel Galuba, Wojciech~Howes, Po-Yao Huang, Li~Shang-Wen, Ishan Misra, Michael Rabbat, Vasu Sharma, Gabriel Synnaeve, Hu~Xu, Herv{\'e} Jegou, Julien Mairal, Patrick Labatut, Armand Joulin, and Piotr Bojanowski. 2023.
\newblock {DINOv2: Learning robust visual features without supervision}.
\newblock \emph{arXiv:2304.07193}.

\bibitem[{Ordonez et~al.(2011{\natexlab{a}})Ordonez, Kulkarni, and Berg}]{ordonez2011sbucaptions}
Vicente Ordonez, Girish Kulkarni, and Tamara Berg. 2011{\natexlab{a}}.
\newblock {Im2Text: Describing Images Using 1 Million Captioned Photographs}.
\newblock In \emph{Advances in Neural Information Processing Systems}, volume~24.

\bibitem[{Ordonez et~al.(2011{\natexlab{b}})Ordonez, Kulkarni, and Berg}]{ordonez2011im2text}
Vicente Ordonez, Girish Kulkarni, and Tamara Berg. 2011{\natexlab{b}}.
\newblock Im2text: Describing images using 1 million captioned photographs.
\newblock \emph{Advances in neural information processing systems}, 24.

\bibitem[{Radford et~al.(2021)Radford, Kim, Hallacy, Ramesh, Goh, Agarwal, Sastry, Askell, Mishkin, Clark, Krueger, and Sutskever}]{radford2021clip}
Alec Radford, Jong~Wook Kim, Chris Hallacy, Aditya Ramesh, Gabriel Goh, Sandhini Agarwal, Girish Sastry, Amanda Askell, Pamela Mishkin, Jack Clark, Gretchen Krueger, and Ilya Sutskever. 2021.
\newblock Learning transferable visual models from natural language supervision.
\newblock In \emph{Proceedings of the 38th International Conference on Machine Learning}, volume 139, pages 8748--8763. PMLR.

\bibitem[{Sharma et~al.(2018{\natexlab{a}})Sharma, Ding, Goodman, and Soricut}]{sharma2018conceptualcaptions}
Piyush Sharma, Nan Ding, Sebastian Goodman, and Radu Soricut. 2018{\natexlab{a}}.
\newblock \href {https://doi.org/10.18653/v1/P18-1238} {Conceptual captions: A cleaned, hypernymed, image alt-text dataset for automatic image captioning}.
\newblock In \emph{Proceedings of the 56th Annual Meeting of the Association for Computational Linguistics (Volume 1: Long Papers)}, pages 2556--2565.

\bibitem[{Sharma et~al.(2018{\natexlab{b}})Sharma, Ding, Goodman, and Soricut}]{sharma-etal-2018-conceptual}
Piyush Sharma, Nan Ding, Sebastian Goodman, and Radu Soricut. 2018{\natexlab{b}}.
\newblock \href {https://doi.org/10.18653/v1/P18-1238} {Conceptual captions: A cleaned, hypernymed, image alt-text dataset for automatic image captioning}.
\newblock In \emph{Proceedings of the 56th Annual Meeting of the Association for Computational Linguistics (Volume 1: Long Papers)}, pages 2556--2565, Melbourne, Australia. Association for Computational Linguistics.

\bibitem[{Su et~al.(2020)Su, Zhu, Cao, Li, Lu, Wei, and Dai}]{su2020vlbert}
Weijie Su, Xizhou Zhu, Yue Cao, Bin Li, Lewei Lu, Furu Wei, and Jifeng Dai. 2020.
\newblock {VL-BERT: Pre-training of Generic Visual-Linguistic Representations}.
\newblock In \emph{International Conference on Learning Representations}.

\bibitem[{Sun et~al.(2019)Sun, Myers, Vondrick, Murphy, and Schmid}]{sun2019videobert}
Chen Sun, Austin Myers, Carl Vondrick, Kevin Murphy, and Cordelia Schmid. 2019.
\newblock {VideoBERT: A Joint Model for Video and Language Representation Learning}.
\newblock In \emph{Proceedings of the IEEE/CVF International Conference on Computer Vision}.

\bibitem[{Tan and Bansal(2019)}]{tan2019lxmert}
Hao Tan and Mohit Bansal. 2019.
\newblock \href {https://doi.org/10.18653/v1/D19-1514} {{LXMERT}: Learning cross-modality encoder representations from transformers}.
\newblock In \emph{Proceedings of the 2019 Conference on Empirical Methods in Natural Language Processing and the 9th International Joint Conference on Natural Language Processing}, pages 5100--5111.

\bibitem[{Tolstikhin et~al.(2021)Tolstikhin, Houlsby, Kolesnikov, Beyer, Zhai, Unterthiner, Yung, Steiner, Keysers, Uszkoreit, Lucic, and Dosovitskiy}]{tolstikhin2021mlp}
Ilya~O Tolstikhin, Neil Houlsby, Alexander Kolesnikov, Lucas Beyer, Xiaohua Zhai, Thomas Unterthiner, Jessica Yung, Andreas Steiner, Daniel Keysers, Jakob Uszkoreit, Mario Lucic, and Alexey Dosovitskiy. 2021.
\newblock {MLP-Mixer: An all-MLP architecture for vision}.
\newblock In \emph{Advances in Neural Information Processing Systems}, volume~34, pages 24261--24272.

\bibitem[{Vouitsis et~al.(2024)Vouitsis, Liu, Gorti, Villecroze, Cresswell, Yu, Loaiza-Ganem, and Volkovs}]{vouitsis2024data}
No{\"e}l Vouitsis, Zhaoyan Liu, Satya~Krishna Gorti, Valentin Villecroze, Jesse~C Cresswell, Guangwei Yu, Gabriel Loaiza-Ganem, and Maksims Volkovs. 2024.
\newblock Data-efficient multimodal fusion on a single gpu.
\newblock In \emph{Proceedings of the IEEE/CVF Conference on Computer Vision and Pattern Recognition}, pages 27239--27251.

\bibitem[{Wang et~al.(2022{\natexlab{a}})Wang, Li, Li, He, Huang, Zhao, Zhang, Xu, Liu, Wang, Xing, Chen, Pan, Yu, Wang, Wang, and Qiao}]{wang2022internvideo}
Yi~Wang, Kunchang Li, Yizhuo Li, Yinan He, Bingkun Huang, Zhiyu Zhao, Hongjie Zhang, Jilan Xu, Yi~Liu, Zun Wang, Sen Xing, Guo Chen, Junting Pan, Jiashuo Yu, Yali Wang, Limin Wang, and Yu~Qiao. 2022{\natexlab{a}}.
\newblock {InternVideo: General video foundation models via generative and discriminative learning}.
\newblock \emph{arXiv:2212.03191}.

\bibitem[{Wang et~al.(2022{\natexlab{b}})Wang, Yu, Yu, Dai, Tsvetkov, and Cao}]{wang2022simvlm}
Zirui Wang, Jiahui Yu, Adams~Wei Yu, Zihang Dai, Yulia Tsvetkov, and Yuan Cao. 2022{\natexlab{b}}.
\newblock Sim{VLM}: Simple visual language model pretraining with weak supervision.
\newblock In \emph{International Conference on Learning Representations}.

\bibitem[{Wu et~al.(2023)Wu, Fei, Qu, Ji, and Chua}]{wu2023next}
Shengqiong Wu, Hao Fei, Leigang Qu, Wei Ji, and Tat-Seng Chua. 2023.
\newblock {NExT-GPT: Any-to-any multimodal LLM}.
\newblock \emph{arXiv:2309.05519}.

\bibitem[{Xiao et~al.(2023)Xiao, Liu, Zhang, and Muennighof}]{xiao2023c}
Shitao Xiao, Zheng Liu, Peitian Zhang, and Niklas Muennighof. 2023.
\newblock C-pack: Packaged resources to advance general chinese embedding.
\newblock \emph{arXiv:2309.07597}.

\bibitem[{Yang et~al.(2021)Yang, Liu, Huang, Wan, Wen, and Guan}]{yang2021infrared}
Yong Yang, Jiaxiang Liu, Shuying Huang, Weiguo Wan, Wenying Wen, and Juwei Guan. 2021.
\newblock Infrared and visible image fusion via texture conditional generative adversarial network.
\newblock \emph{IEEE Transactions on Circuits and Systems for Video Technology}, 31(12):4771--4783.

\bibitem[{Young et~al.(2014{\natexlab{a}})Young, Lai, Hodosh, and Hockenmaier}]{young2014image}
Peter Young, Alice Lai, Micah Hodosh, and Julia Hockenmaier. 2014{\natexlab{a}}.
\newblock From image descriptions to visual denotations: New similarity metrics for semantic inference over event descriptions.
\newblock \emph{Transactions of the Association for Computational Linguistics}, 2:67--78.

\bibitem[{Young et~al.(2014{\natexlab{b}})Young, Lai, Hodosh, and Hockenmaier}]{young2014flickr30k}
Peter Young, Alice Lai, Micah Hodosh, and Julia Hockenmaier. 2014{\natexlab{b}}.
\newblock \href {https://doi.org/10.1162/tacl_a_00166} {From image descriptions to visual denotations: New similarity metrics for semantic inference over event descriptions}.
\newblock \emph{Transactions of the Association for Computational Linguistics}, 2:67--78.

\bibitem[{Yuan et~al.(2021)Yuan, Chen, Chen, Codella, Dai, Gao, Hu, Huang, Li, Li, Liu, Liu, Liu, Lu, Shi, Wang, Wang, Xiao, Xiao, Yang, Zeng, Zhou, and Zhang}]{yuan2021florence}
Lu~Yuan, Dongdong Chen, Yi-Ling Chen, Noel Codella, Xiyang Dai, Jianfeng Gao, Houdong Hu, Xuedong Huang, Boxin Li, Chunyuan Li, Ce~Liu, Mengchen Liu, Zicheng Liu, Yumao Lu, Yu~Shi, Lijuan Wang, Jianfeng Wang, Bin Xiao, Zhen Xiao, Jianwei Yang, Michael Zeng, Luowei Zhou, and Pengchuan Zhang. 2021.
\newblock Florence: A new foundation model for computer vision.
\newblock \emph{arXiv:2111.11432}.

\bibitem[{Zhai et~al.(2022)Zhai, Wang, Mustafa, Steiner, Keysers, Kolesnikov, and Beyer}]{zhai2022lit}
Xiaohua Zhai, Xiao Wang, Basil Mustafa, Andreas Steiner, Daniel Keysers, Alexander Kolesnikov, and Lucas Beyer. 2022.
\newblock {LiT: Zero-Shot Transfer With Locked-Image Text Tuning}.
\newblock In \emph{Proceedings of the IEEE/CVF Conference on Computer Vision and Pattern Recognition}, pages 18123--18133.

\bibitem[{Zhang et~al.(2018)Zhang, Cisse, Dauphin, and Lopez-Paz}]{zhang2018mixup}
Hongyi Zhang, Moustapha Cisse, Yann~N. Dauphin, and David Lopez-Paz. 2018.
\newblock {mixup: Beyond Empirical Risk Minimization}.
\newblock In \emph{International Conference on Learning Representations}.

\bibitem[{Zhang et~al.(2023)Zhang, Gong, Zhang, Li, Qiao, Ouyang, and Yue}]{zhang2023metatransformer}
Yiyuan Zhang, Kaixiong Gong, Kaipeng Zhang, Hongsheng Li, Yu~Qiao, Wanli Ouyang, and Xiangyu Yue. 2023.
\newblock Meta-transformer: A unified framework for multimodal learning.
\newblock \emph{arXiv:2307.10802}.

\end{thebibliography}

\newpage
\appendix
\section{Appendix}

In this section, we present additional implementation details, experiment results, theoretical analysis, pseudo code and supplements. The content structure is outlined as follows:

\begin{itemize}
    \item Section~\ref{app0} - Supplementary Experimental Results
    % \item Section~\ref{gate} - Gate
        \begin{itemize}
        \item Section~\ref{app00} - Results of \textsc{Modest-Align} in Zero-shot Classification
        \item Section~\ref{app01} - Parameter Search for \textsc{Modest-Align}
    \end{itemize}
    \item Section~\ref{app1} - Assessing the Match Quality of Image-text Datasets with CLIP
    % \item Section~\ref{gate} - Gate
        \begin{itemize}
        \item Section~\ref{app11} - Similarity Calculation
        \item Section~\ref{app12} - Similarity in Four Datasets
    \end{itemize}
    \item Section~\ref{ta_es} - Theoretical Analysis for Embedding Smoothing

    \item Section~\ref{ta_rp} - Theoretical Analysis for Random Perturbation
    \item Section~\ref{app4} - Implementation Details
\end{itemize}

\begin{table}[h!]
\centering
\caption{Ablation for N on Different Datasets. To systematically investigate the impact of batch size on performance, we conduct ablation experiments with varying batch sizes ($1k, 2k, 5k, 10k, 15k$). We evaluate how these changes affect \textsc{Modest-Align}’s performance, especially on text-to-image and image retrieval metrics, to better understand the role of batch size in embedding smoothing. Our results indicate that a batch size of 10k offers optimal performance.}
\label{ablation-comparison-N}
\resizebox{1\linewidth}{!}{
\begin{tabular}{c|ccccc} 
\toprule
\textbf{Ablation for N} & {$1k$} & {$2k$} & {$5k$}    & {$10k$}   & {$15k$}  \\ 
\midrule
T$\rightarrow$I (R@1)   & 58.30       & 59.08       & 59.30          & \textbf{60.80} & 59.38         \\
I$\rightarrow$T (R@1)   & 70.00       & 71.70       & \textbf{73.10} & 72.60          & 73.00         \\
Average                 & 64.15       & 65.39       & 66.20          & \textbf{66.70} & 66.19         \\
\bottomrule
\end{tabular}
}
\end{table}

\begin{figure}[htbp]
\centering
% \label{logits_dataset}
\includegraphics[width=0.5\textwidth]{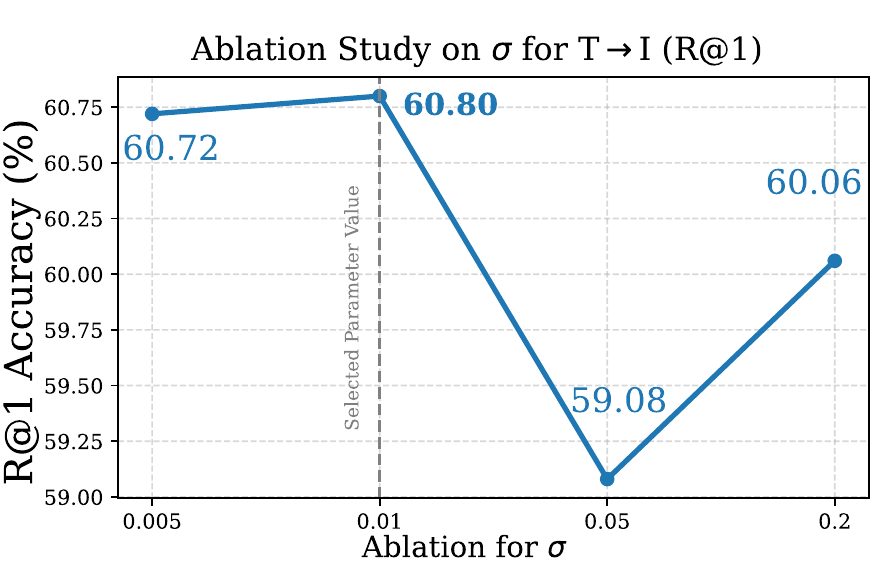}
\caption{Ablation for $\sigma$ on R@1. $\sigma$ Regulates random perturbation intensity, introducing noise to simulate uncertainty in embeddings and reduce overconfidence from weakly correlated data. Smaller $\sigma$
 ensures stability, while larger $\sigma$
 risks excessive uncertainty.}
\label{ablation-sigma}
\end{figure}

\begin{figure}[htbp]
\centering
% \label{logits_dataset}
\includegraphics[width=0.5\textwidth]{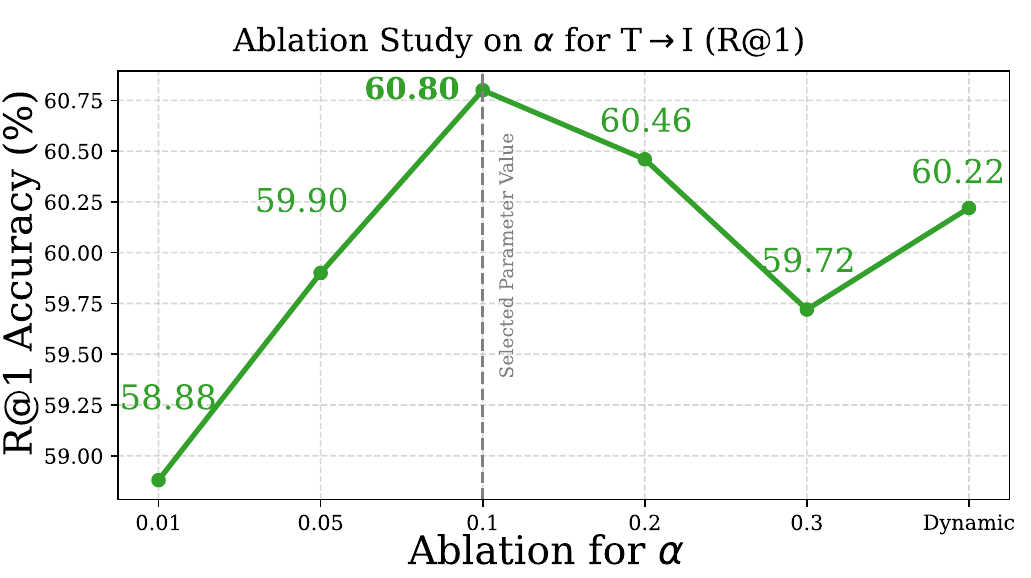}
\caption{Ablation for $\alpha$ on R@1. $\alpha$ Controls embedding smoothing to reduce overconfidence when handling weakly correlated positive samples. It smooths the embedding space, making similarity predictions for image-text pairs more robust. An appropriate $\alpha$ enhances generalization while minimizing reliance on perfectly matched positive samples.}
\label{ablation-alpha}
\end{figure}

\begin{figure}[htbp]
\centering
% \label{logits_dataset}
\includegraphics[width=0.5\textwidth]{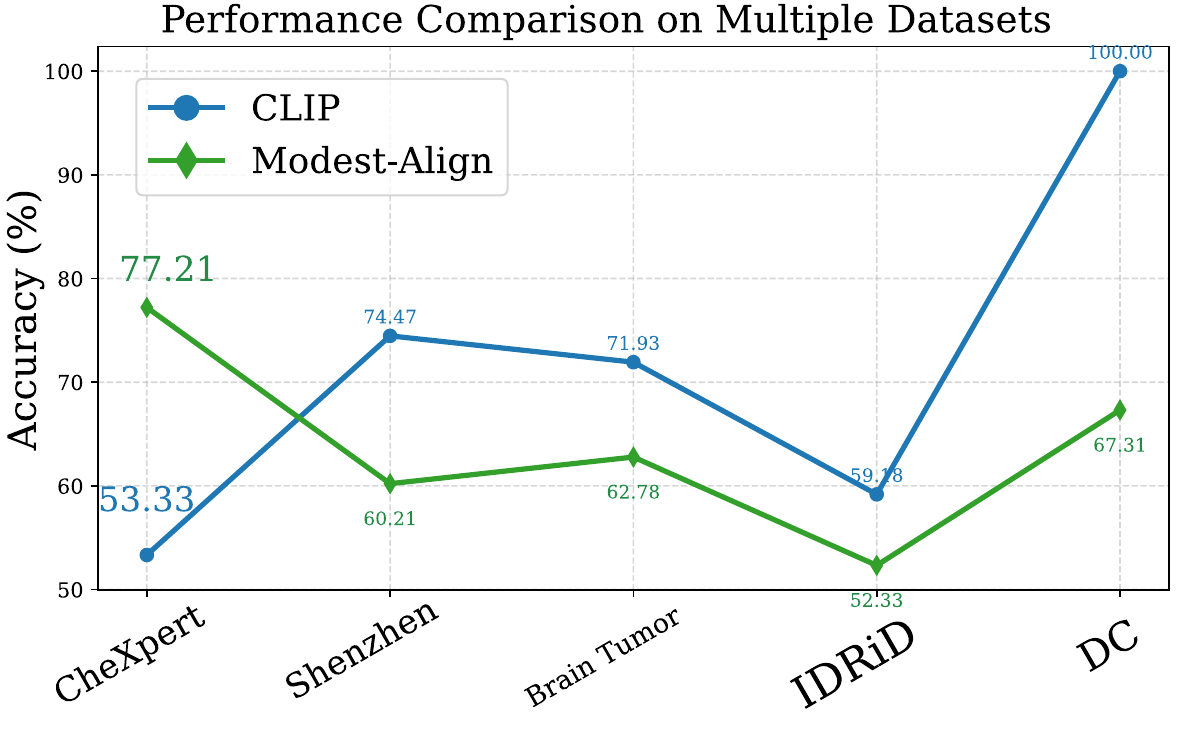}
\caption{Comparison between CLIP and \textsc{Modest-Align} on Various Datasets in Zero-shot Classification}
\label{dataset-comparison-zeroshot}
\end{figure}

\begin{figure}[t]
% \label{logits_dataset}
\centering
\includegraphics[width=0.5\textwidth]{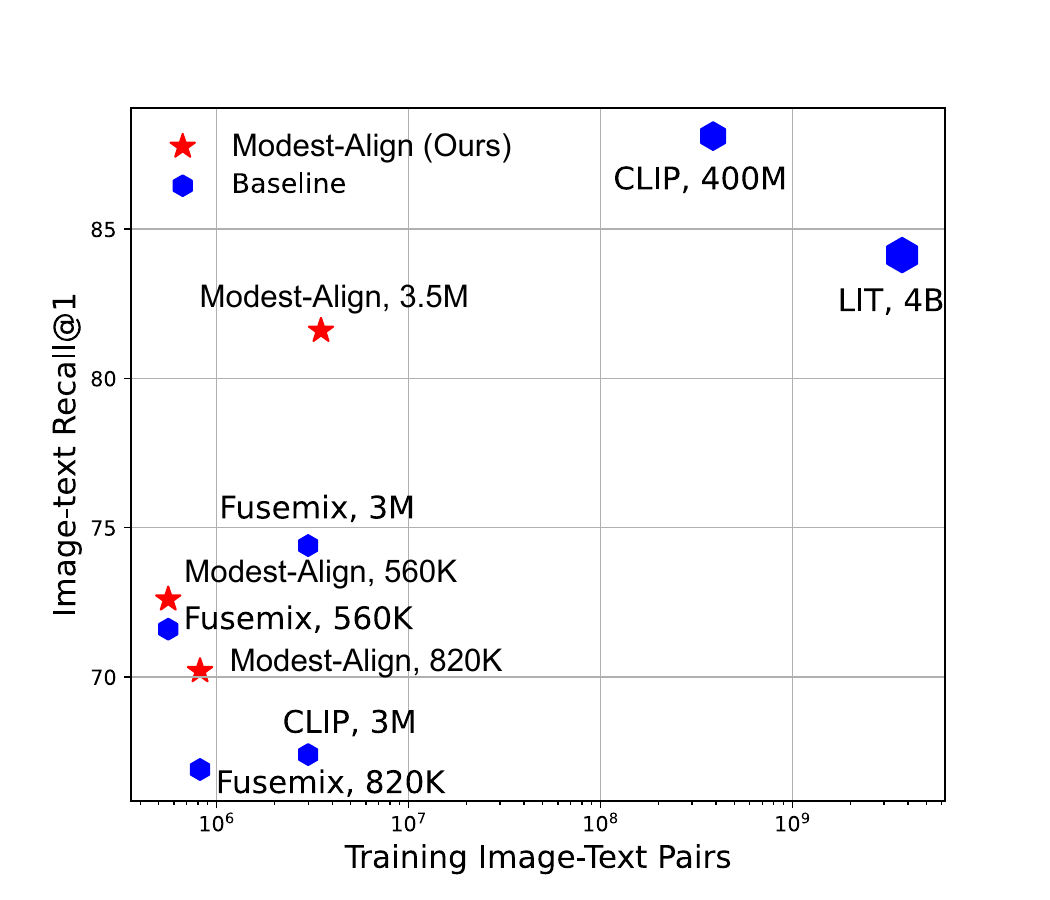}
\caption{
% The performance of text-to-image retrieval is evaluated on the Flickr30K test set \cite{young2014flickr30k}, with the x-axis in log-scale showing the impact of varying numbers of image-text pairs used during training. This highlights how training scale influences retrieval effectiveness.
Image-to-Text retrieval performance on the Flickr30K test set \cite{young2014flickr30k} is plotted against the number of training pairs on a log-scale x-axis, illustrating how training volume impacts effectiveness.
} 
\label{Figure_1}
% \vspace{-2em}
\end{figure}

\subsection{Supplementary Experimental Results}
\label{app0}
\subsubsection{Results of \textsc{Modest-Align} in Zero-shot Classification}
\label{app00}
{\textsc{Modest-Align} aims to optimize cross-modal alignment while minimizing data and resource demands. Evaluating zero-shot classification is essential to further validate its generality.}

To evaluate \textsc{Modest-Align}’s out-of-distribution (OoD) performance on tasks like zero-shot classification and understand its cross-task applicability, we analyzed its performance on diverse medical datasets (CheXpert, Shenzhen, IDRiD, Brain Tumor \cite{liu2023chatgpt}). Specifically, 50 images from different categories were selected as subsets. As shown in the table, \textsc{Modest-Align} achieved a 77.21\% accuracy on the CheXpert, significantly outperforming CLIP’s 53.33\% accuracy. On other datasets, \textsc{Modest-Align}’s results were close to CLIP’s classification performance (\autoref{dataset-comparison-zeroshot}).

To validate \textsc{Modest-Align}’s accuracy on natural images, we tested it on Kaggle’s open-source D\&C dataset. Results showed that CLIP achieved 100.00\% accuracy, significantly outperforming \textsc{Modest-Align}’s 67.31\%. These results highlight CLIP’s strength in its advantage of pretraining on a 400M dataset. However, \textsc{Modest-Align} demonstrated effectiveness across multiple datasets in zero-shot classification tasks, underscoring its generalization capabilities.

With only 3.5M data, \textsc{Modest-Align} achieves performance comparable to or even surpassing CLIP on certain datasets, while using 100 times less data and over 600 times less training time than CLIP, which requires 3000 GPU days and 400M training data. This highlights the importance of \textsc{Modest-Align}’s applicability in resource-constrained environments.

\subsubsection{Parameter Search for \textsc{Modest-Align}}
\label{app01}
To systematically investigate the impact of batch size on performance, we conduct ablation experiments with varying batch sizes (1k, 2k, 5k, 10k, and 15k). We evaluate how these changes affect \textsc{Modest-Align}’s performance, especially on text-to-image and image retrieval metrics, to better understand the role of batch size in embedding smoothing. Our results indicate that a batch size of 10k offers optimal performance (\autoref{ablation-comparison-N}).

Impact of parameter $\sigma$: We fix $\lambda=1$ and treat $\sigma$ as the sole hyper-parameter in Equation~7. As the control parameter for random perturbation intensity, $\sigma$ influences the model’s robustness in the embedding space. Experimental results indicate:
Smaller $\sigma$ values (e.g., 0.005) maintain embedding stability but may be insufficient for handling weakly correlated samples. Larger $\sigma$ values (e.g., 0.05 or higher) enhance the model’s adaptability to noise but may overly disturb the feature space, negatively impacting alignment accuracy (\autoref{ablation-sigma}).

Impact of parameter $\alpha$: As the control parameter for embedding smoothing, $\alpha$ influences the distribution smoothness between positive and negative pairs. An appropriate 
$\alpha$ value effectively mitigates overconfidence issues and enhances generalization:
Lower $\alpha$ values (e.g., 0.05) may fail to adequately reduce the model’s overconfidence. Higher $\alpha$ values (e.g., 0.3) may make the model less confident even for fully matched positive samples, causing interference with positive pair alignment (\autoref{ablation-alpha}).

\begin{figure*}[t]
\centering
\label{fig:ambiguous_samples}
\includegraphics[width=\textwidth]{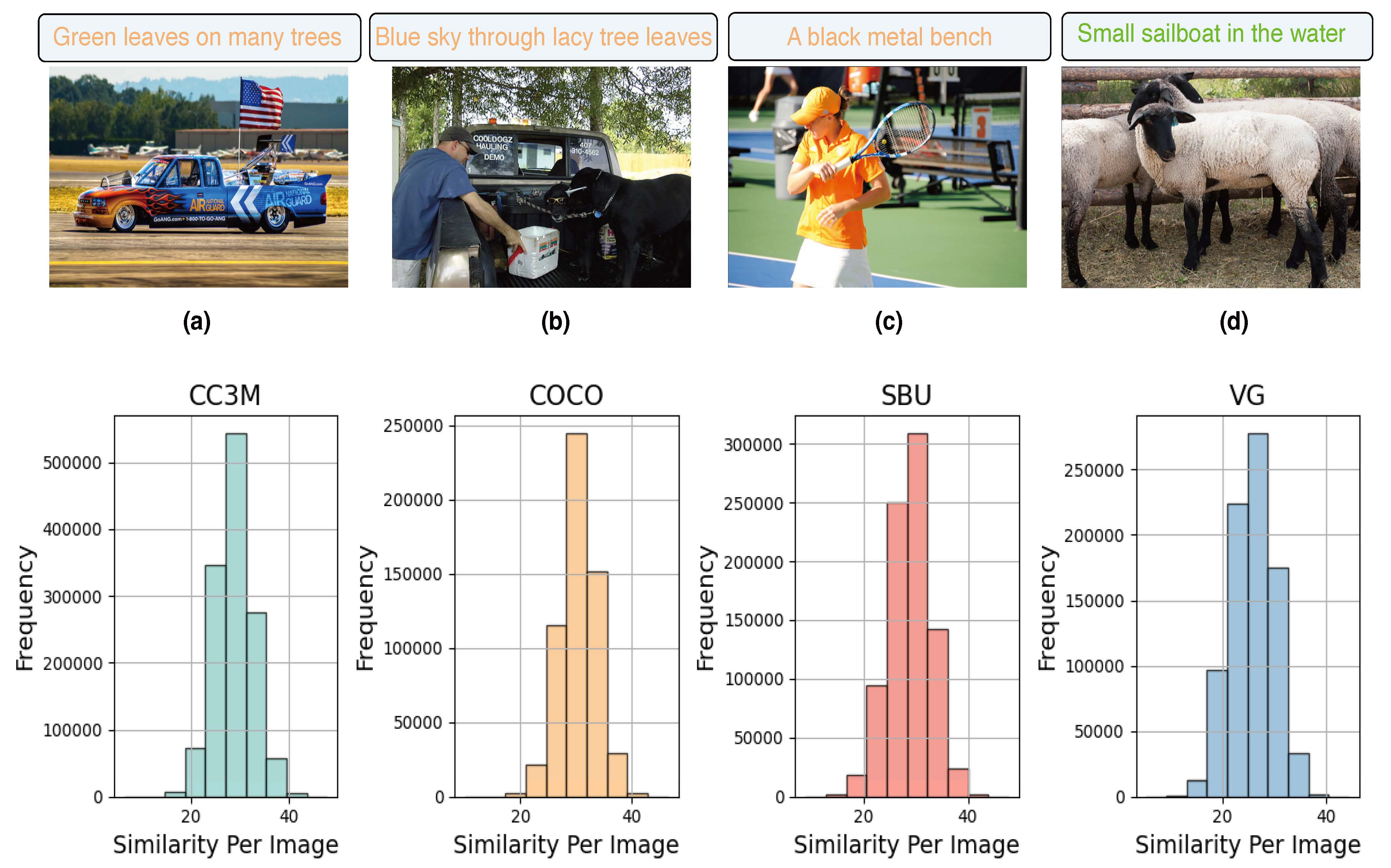}
\caption{
The upper figure mentions ''ambiguous samples" in datasets, including partially matched samples (a, b, c) and completely unmatched samples (d). During contrastive learning training, it can lead to positive pairs not being truly positive. The lower figure use CLIP to compute similarity for four datasets (For details, see \autoref{tab:Distribution of logits}), revealing generally moderate alignment between images and texts (mostly ranging from 20-40\%). Notably, the COCO dataset shows higher alignment (indicative of superior data quality), which correlates with relatively higher performance metrics.
} 
\label{Figure_datasets}
\end{figure*}

\subsection{Assessing the Match Quality of Image-text Datasets with CLIP}
\label{app1}

\subsubsection{Similarity Calculation}
\label{app11}

In the context of contrastive learning models such as CLIP, the similarity produced for a given image-text pair is closely related to the cosine similarity of their respective embeddings, modulated by a temperature scaling factor \( \tau \). Specifically, let \(Similarity(i, j) \) denote the similarity score for image \( i \) and text \( j \). This score can be mathematically expressed as:

\scriptsize
\begin{equation}
   Similarity(i, j) = \frac{\text{image\_embedding}(i) \cdot \text{text\_embedding}(j)}{\|\text{image\_embedding}(i)\| \|\text{text\_embedding}(j)\|} \times \frac{1}{\tau},
   \tag{1}
\end{equation}

\normalsize 

where \( \tau \) is the temperature coefficient, which is set to \( 0.01 \). 
% The cosine similarity term ranges from -1 to 1. 
Consequently, when multiplied by the temperature coefficient \( \frac{1}{\tau} \), the similarity score will be constrained within the new range.

\begin{table*}[htbp]
\centering
\caption{The performance of \textsc{Modest-Align} and Fusemix \cite{vouitsis2024data} on Flickr30K and MS-COCO test sets. By evaluating on multiple datasets with varying sizes and complexities, we show that our \textsc{Modest-Align} model exhibits strong generalization capabilities and achieves SoTA performance on image retrieval tasks. Bold signifies the best.}
\label{experiment-table1-1}
\resizebox{\textwidth}{!}{
\begin{tabular}{cccccccc|cccccc}
\toprule
                                                                      &                          & \multicolumn{6}{c|}{Flickr30K (1K test set)}                                                                           & \multicolumn{6}{c}{MS-COCO (5K test set)}                                                                             \\ \cmidrule{3-14} 
                                                                      &                          & \multicolumn{3}{c}{$\text{text} \rightarrow \text{image}$} & \multicolumn{3}{c|}{$\text{image} \rightarrow \text{text}$} & \multicolumn{3}{c}{$\text{text} \rightarrow \text{image}$} & \multicolumn{3}{c}{$\text{image} \rightarrow \text{text}$} \\
\multirow{-3}{*}{Training Dataset Size}                               & \multirow{-3}{*}{Method} & R@1               & R@5               & R@10              & R@1                & R@5               & R@10              & R@1               & R@5               & R@10              & R@1               & R@5               & R@10              \\ \midrule
\cellcolor[HTML]{f5f5f5}                                              & Fusemix                  & 57.80             & 83.38             & 89.54             & 71.60              & 91.10             & 95.00             & 43.68             & 72.53             & 82.53             & 57.22             & 83.40             & 90.68             \\
\multirow{-2}{*}{\cellcolor[HTML]{f5f5f5}$\approx 560K$}                        & \textsc{Modest-Align} (ours)           & \textbf{60.80}             & \textbf{84.82}             & \textbf{90.82}             & \textbf{72.60}              & \textbf{93.30}             & \textbf{95.70}             & \textbf{45.11}             & \textbf{74.16}             & \textbf{83.05}             & \textbf{57.76}             & \textbf{84.38}             & \textbf{91.60}             \\ \midrule
\cellcolor[HTML]{e0e0e0}                                              & Fusemix                  & 43.32             & 72.48             & 81.36             & 61.60              & 86.30             & 92.10             & 22.98             & 47.64             & 59.37             & \textbf{32.36}             & 56.34             & \textbf{67.50}             \\
\multirow{-2}{*}{\cellcolor[HTML]{e0e0e0}$\approx 840K$}                        & \textsc{Modest-Align} (ours)           & \textbf{47.80}             & \textbf{75.94}             & \textbf{84.18}             & \textbf{62.10}              & \textbf{87.00}             & \textbf{92.30 }            & \textbf{24.35}             & \textbf{49.00}             & \textbf{60.71}             & 31.66             & \textbf{56.54}             & 67.28             \\ \midrule
\cellcolor[HTML]{c4c4c4}                                              & Fusemix                  & 61.40             & 84.82             & 90.46             & 77.20              & 94.40             & 97.20             & 43.76             & 72.74             & 82.36             & 60.94             & 84.72             & 91.68             \\
\multirow{-2}{*}{\cellcolor[HTML]{c4c4c4}$\approx 2M$}                          & \textsc{Modest-Align} (ours)           & \textbf{62.10}             & \textbf{85.52}             & \textbf{90.76}             & \textbf{77.80}              & \textbf{94.60}             & \textbf{97.70}             & \textbf{44.27}             & \textbf{72.93}             & \textbf{82.77}             & \textbf{61.42}             & \textbf{84.98}             & \textbf{91.92}             \\ \midrule
\cellcolor[HTML]{a8a8a8}                                              & Fusemix                  & 64.28             & 87.60             & 91.86             & 81.20              & \textbf{96.40}             & 98.20             & 45.54             & 73.35             & 82.73             & 62.56             & \textbf{86.38}             & \textbf{93.04}             \\
\multirow{-2}{*}{\cellcolor[HTML]{a8a8a8}$\approx 3.5M$}                          & \textsc{Modest-Align} (ours)           & \textbf{65.72}             & \textbf{87.82}             & \textbf{92.50}             & \textbf{81.60}              & 96.00             & \textbf{98.30}             & \textbf{45.86}             & \textbf{73.73}             & \textbf{83.16}             & \textbf{62.86}             & 85.66             & 92.42             \\ \bottomrule
\end{tabular}
}
\vspace{-1em}
\end{table*}

\subsubsection{Similarity in Four Datasets}
\label{app12}
% To provide a Match Quality of Image-text Datasets, we conducted experiments on four widely used image-text datasets: COCO, CC3M, SBU, and VG.
To assess the match quality of image-text datasets in the main text, we conducted similarity calculation experiments on four widely used image-text datasets: COCO, CC3M, SBU, and VG.
These datasets cover a diverse range of image content, from everyday objects to complex scenes, and vary in terms of size and annotation quality. 
\autoref{tab:Distribution of logits} presents a summary of these datasets, detailing the number of image-text pairs in each and the ranges of similarity scores calculated for different image-text pairings.

In a similarity analysis, the COCO dataset exhibited the highest mean similarity per image at 30.48, indicating a strong and consistent association between images and text. The CC3M and SBU datasets demonstrated similar mean similarities of 28.80, reflecting comparable levels of alignment quality. This highlights how dataset structure and complexity significantly influence model performance in multimodal tasks, as shown in \autoref{Figure_datasets} (b).
\autoref{tab:Distribution of logits} shows the distribution of similarity for Image-Text pairs across four datasets. It can be observed that the values are concentrated around the 30 range, indicating that when CLIP is used as a scoring model, the resulting similarity tends to cluster within a relatively modest range.

\begin{table*}[htbp]
    \centering
    \caption{
    Similarity Per Image-Text Across Ranges for Different Datasets. 
    The table shows the count of similarity scores within specific similarity ranges for each dataset (CC3M, COCO, SBU, and VG). Each row corresponds to a dataset, and each column represents a range of similarity values.
    }
    \label{tab:Distribution of logits}
    \resizebox{\linewidth}{!}{ 
    \centering
    \begin{tabular}{lccccccccc}
        \toprule
        & \textbf{[5,10)} & \textbf{[10, 15)} & \textbf{[15, 20)} & \textbf{[20, 25)} & \textbf{[25, 30)} & \textbf{[30, 35)} & \textbf{[35, 40)} & \textbf{[40, 45)} & \textbf{[45, 50)} \\
        \midrule
        \textbf{CC3M} & 11  & 400   & 14877  & 189355 & 616912 & 408777 & 70836  & 3389  & 32 \\
        \textbf{COCO}     & 1   & 29    & 1187   & 27566  & 216780 & 275238 & 44607  & 1331  & 8  \\
        \textbf{SBU}      & 15  & 647   & 15268  & 127793 & 365869 & 284130 & 45279  & 1702  & 24 \\
        \textbf{VG}       & 39  & 3149  & 68441  & 267798 & 342860 & 132259 & 7151   & 77    & 0  \\
        \bottomrule
    \end{tabular}
    } 
\end{table*}

\begin{figure}[t!]
\centering
\label{Days}
\includegraphics[width=0.5\textwidth]{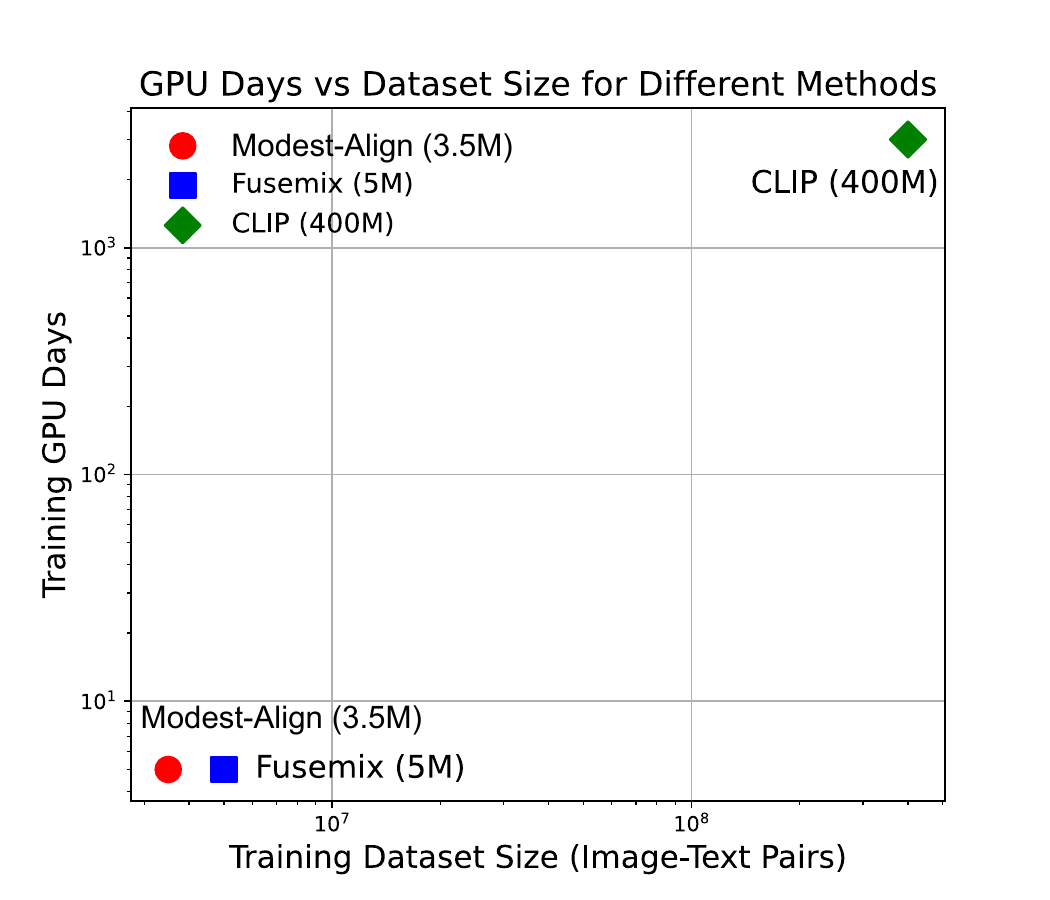}
\caption{
In terms of training efficiency, CLIP requires 3000 GPU days for training on 400 million data pairs, while \textsc{Modest-Align} needs only approximately 5 GPU days for 3.5 million data pairs. Our proposed \textsc{Modest-Align} aims to enhance computational and data efficiency for modal alignment in resource-efficient settings. It has outperformed state-of-the-art methods on public datasets, notably improving the R@1 score in retrieval tasks. With only 3.5M data, \textsc{Modest-Align} exceeds CLIP’s performance on the MS COCO dataset and does so with 100 times less data and over 600 times less training time than CLIP, which requires 3000 GPU days and 400M training data.
} 
\label{Figure_days}
\vspace{-1em}
\end{figure}

\subsection{Theoretical Analysis for Embedding Smoothing}
\label{ta_es}

Embedding Smoothing effectively increases the entropy of the target distributions by assigning non-zero probabilities to all classes (examples in the batch). This reduction in confidence prevents the model from becoming overly reliant on specific training examples, thereby enhancing its robustness. The inclusion of the smoothing parameter \( \alpha \) allows for control over the degree of smoothing applied, enabling a balance between model confidence and generalization ability.
To provide a theoretical understanding of how Embedding Smoothing improves generalization, we analyze its impact on the loss function and the model's predictions.

In standard contrastive learning without smoothing, the loss for a positive pair is:

\small
\begin{equation}
    \mathcal{L}_{\text{pos}} = -\log \frac{\exp\left( \text{sim}\left( A_X(z_x), A_Y(z_y) \right) / \tau \right)}{\sum_{j=1}^{N} \exp\left( \text{sim}\left( A_X(z_x), A_Y(z_j) \right) / \tau \right)},
\end{equation}
\normalsize

This loss encourages the model to maximize the similarity between positive pairs and minimize it between negative pairs. However, it can lead to overconfident predictions, as the model focuses heavily on the positive pair.
With Embedding Smoothing, the loss incorporates the smoothed target distribution \( \tilde{y} \), and the KL divergence becomes:

\begin{equation}
    \mathcal{L}\left(A_X(z_x), \tilde{Y}; Z_Y\right) = - \sum_{i=1}^{N} \tilde{y}_i \log p_i,
\end{equation}

where \( p_i \) is the predicted probability for the \( i \)-th example in the batch:

\begin{equation}
    p_i = \frac{\exp\left( \text{sim}\left( A_X(z_x), A_Y(z_i) \right) / \tau \right)}{\sum_{j=1}^{N} \exp\left( \text{sim}\left( A_X(z_x), A_Y(z_j) \right) / \tau \right)},
\end{equation}

By assigning non-zero probabilities \( \tilde{y}_i \) to all classes, the loss function penalizes the model not only for the positive pair but also for negative pairs, albeit to a lesser extent. This encourages the model to produce a probability distribution that is more uniform and less confident.

% Entropy Increase and Generalization

Analyzing from the perspective of information entropy: The entropy \( H(\tilde{y}) \) of the smoothed target distribution is higher than that of a one-hot distribution. The entropy of \( \tilde{y} \) is:
\scriptsize
\begin{equation}
    H(\tilde{y}) = - \left( (1 - \alpha) \log (1 - \alpha) + (N - 1) \left( \frac{\alpha}{N - 1} \log \frac{\alpha}{N - 1} \right) \right),
\end{equation}
\normalsize

Higher entropy in the target distribution leads to smoother gradients during training, which can prevent the model from fitting noise in the training data. This smoothing effect acts as a form of regularization, reducing overfitting.

\textbf{Reduction in Overconfident Predictions:} Embedding Smoothing reduces the Kullback-Leibler divergence between the predicted distribution \( p \) and the uniform distribution \( u \), where \( u_i = \frac{1}{N} \):

\begin{equation}
    \text{KL}(u \| p) = \sum_{i=1}^{N} u_i \log \frac{u_i}{p_i},
\end{equation}

By making \( p \) closer to \( \tilde{y} \), which has higher entropy, the model's predictions become less confident. This can be beneficial because overconfident predictions on training data often lead to poor generalization on unseen data.

\textbf{Connection to Label Smoothing Theory:}
Embedding Smoothing in our context is analogous to label smoothing in classification tasks. Previous works have shown that label smoothing has the following effects:
1). Margin Maximization: It implicitly increases the decision margin between classes, which can improve generalization.
2). Penalization of Confident Wrong Predictions: By smoothing the targets, the loss function penalizes overconfident incorrect predictions more heavily.

Besides, consider the gradient of the loss with respect to the logits \( z \):

\begin{equation}
    \frac{\partial \mathcal{L}}{\partial z_i} = p_i - \tilde{y}_i,
\end{equation}

When using Embedding Smoothing, \( \tilde{y}_i \) is never exactly 0 or 1. This means that the gradients are non-zero for all classes, encouraging the model to adjust its predictions across all examples in the batch. This leads to more generalized feature representations.
By incorporating Embedding Smoothing into the loss function, we introduce a regularization effect that enhances the model's generalization capabilities. The smoothing parameter \( \alpha \) provides a mechanism to control this effect, allowing for a trade-off between fitting the training data and maintaining robustness to unseen data. This theoretical understanding aligns with our experimental observations, where models trained with Embedding Smoothing demonstrate improved performance on validation datasets.

\subsection{Theoretical Analysis for Random Perturbation}
\label{ta_rp}

\textbf{Why Choose Gaussian noise?:}
This paper introduces Gaussian noise as a perturbation in \textsc{Modest-Align} for several reasons:
1).	Well-Defined Mathematical Properties: Gaussian distribution exhibits continuous and smooth probability density functions across the real number line, facilitating theoretical analysis and calculations.
2).	Zero-Mean Symmetry: By choosing a Gaussian distribution with a mean of zero, the added noise is symmetrically balanced around zero, introducing no systematic bias and only increasing the variance, thereby preserving the expected value of embeddings.
3).	Adjustable Perturbation Intensity: The standard deviation of the Gaussian distribution can be precisely controlled, allowing for careful calibration of noise intensity. This flexibility is crucial for introducing an appropriate level of uncertainty to enhance model robustness.
4).	Alignment with Natural Phenomena: According to the Central Limit Theorem, the sum of many independent random variables tends toward a Gaussian distribution. Thus, Gaussian noise effectively simulates random disturbances or measurement errors prevalent in natural and engineering contexts.
5).	Facilitation of Optimization and Training: In deep learning, incorporating Gaussian noise helps to smooth the loss function landscape, avoiding local minima and promoting more effective training processes.
% 6).	Theoretical and Practical Ubiquity: Gaussian distribution is extensively used in machine learning and statistics. Leveraging noise based on this distribution allows easy integration with existing theoretical frameworks and algorithms.

In summary, choosing Gaussian noise as the source of perturbation provides theoretical soundness and practical convenience, aiding in the development of more robust feature representations, preventing overfitting, and enhancing the generalization capabilities of models.

\textbf{Loss Function with Perturbed Embeddings:}
The perturbed embeddings are used in the training loss, specifically in contrastive learning with the InfoNCE loss. The objective is to maximize the similarity between the noisy visual and textual embeddings:

\begin{equation}
    \mathcal{L}_{\text{NCE}} = - \log \frac{\exp(\text{sim}(\tilde{\mathbf{z}}_v, \tilde{\mathbf{z}}_t) / \tau)}{\sum_{j=1}^{N} \exp(\text{sim}(\tilde{\mathbf{z}}_v, \tilde{\mathbf{z}}_t^j) / \tau)},
\end{equation}

where $\text{sim}(\cdot, \cdot)$ represents a similarity function, $\tau$ is a temperature parameter, and $N$ is the number of negative samples. This formulation ensures that the model learns representations that are invariant to noise, thus improving generalization.

\textbf{Noise and Regularization Effect:}
To understand the impact of Gaussian noise on regularization, we first look at the expected value of the perturbed embeddings. Since the noise is zero-mean, the expected value of the perturbed embeddings is identical to the original embeddings:

\begin{equation}
    \mathbb{E}[\tilde{\mathbf{z}}_v] = \mathbb{E}[\mathbf{z}_v + \sigma \epsilon_v] = \mathbf{z}_v,
\end{equation}

However, the variance of the perturbed embeddings increases due to the added noise. The variance of the perturbed embeddings can be calculated as:

\begin{equation}
    \text{Var}[\tilde{\mathbf{z}}_v] = \text{Var}[\mathbf{z}_v + \sigma \epsilon_v] = \text{Var}[\mathbf{z}_v] + \sigma^2 \text{Var}[\epsilon_v],
\end{equation}

Given that $\text{Var}[\epsilon_v] = I$, where $I$ is the identity matrix, the total variance of the perturbed embeddings becomes:

\begin{equation}
    \text{Var}_{\text{total}} = \text{Var}[\mathbf{z}_v] + \sigma^2 I,
\end{equation}

The additional term $\sigma^2 I$ acts as a regularizer, which spreads out the embeddings and prevents the model from becoming overconfident in its predictions.

\textbf{Minimizing the Generalization Error:}
The added noise effectively smooths the decision boundary of the model, which reduces overfitting. By introducing noise, we minimize the generalization error. Assuming the model's prediction function is $f(\mathbf{z})$ and the true function is $f^*(\mathbf{z})$, the goal is to minimize the expected generalization error:
\[
    \mathbb{E}_{\mathbf{z}}[(f(\mathbf{z}) - f^*(\mathbf{z}))^2],
\]
With noise perturbation, the variance in the embeddings increases, which forces the model to learn smoother decision boundaries. The regularization effect introduced by the noise helps bind the generalization error:

\begin{equation}
    \mathbb{E}_{\mathbf{z}}[(f(\mathbf{z}) - f^*(\mathbf{z}))^2] \leq \text{Var}_{\text{total}} = \text{Var}[\mathbf{z}_v] + \sigma^2,
\end{equation}

Thus, the noise helps control the generalization error by ensuring that the model does not overfit to specific features of the training data, which is especially important in cases where the training data contains noise or is limited in size.

\textbf{Noise-Induced Gradient Regularization:}
We can also analyze the effect of noise on the gradient of the loss function. Given a loss function $\mathcal{L}(\mathbf{z}_v, \mathbf{z}_t)$, the gradient with respect to the perturbed embeddings can be expressed as:

\begin{equation}
    \nabla_{\tilde{\mathbf{z}}_v} \mathcal{L}(\tilde{\mathbf{z}}_v, \tilde{\mathbf{z}}_t) = \nabla_{\mathbf{z}_v} \mathcal{L}(\mathbf{z}_v, \mathbf{z}_t) + \sigma \cdot \nabla_{\epsilon_v} \mathcal{L}(\tilde{\mathbf{z}}_v, \tilde{\mathbf{z}}_t),
\end{equation}

The second term, $\sigma \cdot \nabla_{\epsilon_v} \mathcal{L}(\tilde{\mathbf{z}}_v, \tilde{\mathbf{z}}_t)$, acts as a regularizer that prevents the gradient from becoming too large. The gradient is smoothed by the presence of noise, which further prevents overfitting and encourages the model to learn more generalizable patterns.

\textbf{Inference Phase:}
During the inference phase, we remove the Gaussian noise to ensure accurate predictions on unseen data. The embeddings revert to their original clean form:

\begin{equation}
    \mathbf{z}_v = F(\mathbf{x}_v), \quad \mathbf{z}_t = G(\mathbf{x}_t),
\end{equation}

Without the added noise, the model makes precise predictions based on the robust features it learned during training.
Therefore, by adding Gaussian noise to the embeddings during training, we introduce a form of regularization that improves the generalization ability of the model. The noise prevents overfitting by increasing the variance of the embeddings, ensuring that the model learns smoother decision boundaries. This leads to better performance on unseen data and helps minimize the generalization error. The noise-induced gradient regularization further contributes to preventing the model from overfitting to the training data, making it more robust in real-world applications.

\subsection{Implementation Details}
\label{app4}
\label{sec:appendix-impl}
For all experiments, we use the AdamW \cite{loshchilov2018fixing} optimizer during training. We perform learning rate warmup by linearly increasing the learning rate from $10^{-6}$ to $10^{-3}$.
% during the first epoch. 
We then decay the learning rate using a cosine
schedule \cite{loshchilov2016sgdr}.
% We also set our FuseMix Beta distribution hyperparameter as $\alpha=1$ so that the interpolation coefficient is sampled as $\lambda \sim \mathcal{B}(1,1)$.\footnote{$\mathcal{B}(\alpha,\alpha)$ is the uniform distribution when $\alpha=1$, concentrates around $0$ and $1$ when $\alpha<1$, and is unimodal when $\alpha>1$.} We note that when mixup is performed on ambient space, it is common to select small $\alpha$ \cite{zhang2018mixup, verma2021dacl, hao2023mixgen}. 
% This ensures that inputs are only slightly  perturbed so that they remain semantically meaningful. Conversely, in FuseMix, we are operating on the latent space of pre-trained unimodal encoders where we find that relatively larger $\alpha$ can improve performance in our experiments, which suggests that larger perturbations on latent space can remain semantically meaningful (see result in Appendix \ref{sec:appendix-albations}). We next describe specific details and hyperparameters for each task we consider:
We use a depth of 4 for both V-L adapters which we train for 500 epochs with a batch size of $Batch=10$K. We set the learning rate as  \texttt{lr}$=10^{-3}$ and use weight decay of 0.1 during optimization. Additionally, $\tau_t$ is set to 0.07.
The image encoder is DINOv2 ViT-G/14, and for the text side, the text encoder is the BGE large version.
To evaluate the effectiveness of the \textsc{Modest-Align} method for the task of modality alignment, we conducted extensive comparative experiments against SoTA methods across various datasets. 
These datasets include COCO \cite{lin2014microsoft}, VG  \cite{krishna2017visual}, SBU \cite{ordonez2011im2text}, and CC3M \cite{sharma-etal-2018-conceptual}. 
Table \ref{tab:Distribution of logits} provides detailed information about these four datasets. Additionally, we compared different schemes such as Fusemix, CLIP, and LIT. Utilizing a single NVIDIA 3090 GPU for training, \textsc{Modest-Align} demonstrated SoTA performance across datasets of varying sizes.

\end{document}